\documentclass[runningheads]{llncs}

\usepackage[final]{eccv}

\usepackage{eccvabbrv}

\usepackage{graphicx}
\usepackage{booktabs}

\usepackage[accsupp]{axessibility}  

\usepackage{hyperref}

\usepackage{orcidlink}

\usepackage{url}

\usepackage{xcolor}         
\usepackage{graphicx} 
\usepackage{algorithm}
\usepackage{algpseudocode}
\usepackage{listings}
\usepackage{color}
\usepackage{xcolor}

\usepackage{algorithm}
\usepackage{algpseudocode}
\usepackage{listings}
\usepackage{float}
\usepackage{color}
\usepackage{xcolor}

\newcommand*{\samethanks}[1][\value{footnote}]{\footnotemark[#1]}

\begin{document}

\title{PIC: Revisiting INR for Image Coding with Fast Encoding and Sub-Millisecond Decoding} 

\titlerunning{PIC}

\author{
  Xiang Liu\inst{1,3}\thanks{Equal contribution.}\orcidlink{0009-0002-7929-5501} \and
  Jinxiang Wang\inst{1}\samethanks\and
  Bin Chen\inst{2}\thanks{Corresponding author.}\orcidlink{0000-0002-4798-230X} \and 
  Zimo Liu\inst{3}\orcidlink{0009-0000-4014-1543} \and
  Mingyao Hong\inst{3}\orcidlink{0000-0003-1537-6468} \and
  Jiawei Li\inst{4}\orcidlink{0000-0003-3873-8003} \and
  Yaowei Wang \inst{2,3}\orcidlink{0000-0002-6110-4036} \and
  Shu-tao Xia \inst{1,3}\orcidlink{0000-0002-8639-982X}
}

\authorrunning{X. Liu et al.}


\institute{Tsinghua University, China \and
Harbin Institute of Technology, Shenzhen, China  \and
Peng Cheng Laboratory, China  \and 
Joy Future Academy, JD Group, China }

\maketitle

\begin{abstract}

  Implicit neural representation (INR) has achieved remarkable progress in novel view synthesis and image/video coding in recent years. 
  Compared to conventional end-to-end image codecs, INR-based compressors demonstrate significant advantages in decoding complexity. However, their practical application has been hindered by the inferior encoding speed and underutilized decoding efficiency. 
  In this work, we propose a feedforward INR image coding architecture, \textbf{P}ractical \textbf{I}NR Image \textbf{C}odec (PIC), that computes all the necessary information for INR network in a single forward pass, achieving an encoding speed of 20 FPS. Additionally, we implement a highly optimized decoder that reaches 2000 FPS decoding speed, significantly surpassing JPEG's performance at comparable rate-distortion (RD) performance. To the best of our knowledge, this work presents the first learning-based image codec that simultaneously outperforms or is comparable with JPEG in both RD performance and decoding speed while maintaining practical encoding speed. Code is available at \url{https://github.com/actcwlf/PIC}.

\end{abstract}

\section{Introduction}

Image compression has long been a fundamental topic in signal processing and remains a critical technology underpinning more complex systems such as video coding. With the rapid expansion of Internet data, industrial data, AIGC data, and other forms of data, compression technology remains a critically important research area. Image compression technology has undergone significant evolution, encompassing traditional encoders such as JPEG \cite{wallace1992jpeg}, end-to-end models \cite{balle2017end, balle2018variational}, and recently emerged compression methods utilizing implicit neural representation (INR) \cite{dupont2021coin, dupont2022coin++, ladune2022cool} or Gaussian Splatting (GS) \cite{kerbl20233d, zhang2024gaussianimage,li2026gaussianimage++}. Each of these approaches demonstrates distinct strengths and limitations in different aspects of image compression tasks.
Traditional image encoders typically leverage human understanding of signals and visual perception, employing predefined rules to discard visually insignificant information for compression. For instance, JPEG exploits the human eye's reduced sensitivity to certain high-frequency components by decomposing the image into different frequencies using Discrete Cosine Transform (DCT) and selectively retaining the most perceptually critical parts, thereby achieving highly efficient compression. Additionally, its codec design strikes a balance between rate-distortion (RD) performance and encoding/decoding speed. Subsequent traditional encoders further improved RD performance by adopting more sophisticated transformation techniques, such as wavelet transforms \cite{skodras2001jpeg} and predictive coding \cite{bpg}, although at the cost of increased computational complexity and slower processing speeds.

Classic end-to-end image compression algorithms \cite{balle2017end, balle2018variational} conceptually follow the transform coding paradigm used in traditional image codecs. The key difference is that traditional compressors incorporate numerous manually designed transforming rules, whereas the end-to-end approach employs data-driven method to learn the transformation. 
Taking advantage of the powerful learning capability of neural networks, this data-driven nonlinear transformation can effectively model the image distribution prior to the data, thereby achieving RD performance that surpasses traditional image codecs \cite{he2022elic, jiang2023mlic}. Furthermore, significant progress has been made based on generative models \cite{mentzer2020high}, particularly in low bit-rate compression scenarios oriented to perceptual metrics \cite{xiadiffpc}. 


More recently, representation-based methods---such as INRs \cite{dupont2021coin} and 3D Gaussian Splatting \cite{kerbl20233d}---have emerged as promising alternatives. Their key advantage lies in extremely low decoding complexity \cite{ladune2022cool, liu2024efficient}, often achieving orders-of-magnitude speedups over neural codecs \cite{zhang2024gaussianimage}. Some variants also rival traditional codecs in rate-distortion performance \cite{kim2024c3}. However, these gains come at the cost of extremely slow encoding, often requiring minutes or hours to train a model per image, severely limiting real-world deployment.

\begin{table}[t]
    \centering
    \caption{Comparision of different paradigm. E2E and RM are abbreviation for end-to-end and representation model respectively. Feedforward means the method is able to encode an image in one forward pass, instead of a full training process. RD represents rate distortion performance. We selected three representative methods, Factorized \cite{balle2017end}, Hyperprior \cite{balle2018variational}, COIN \cite{dupont2021coin}, Cool-Chic v4.2 (fast)~\cite{ladune2022cool, leguay2023low, kim2024c3} and GaussianImage \cite{zhang2024gaussianimage}, for comparison. BD-Rate are calcualted relative to JPEG on Kodak dataset. Other results are representative value from the same experiment. Due to the lack of a fair FLOPs calculation method, the complexity of GaussianImage is left blank. Detailed numerical results of all metrics are shown in \cref{fig:main_results} and \cref{fig:main_results:clic}.}
    \resizebox{0.99\linewidth}{!}{
\begin{tabular}{ccccccc}
        \toprule
Model & Paradigm & Feedforward & BD-Rate$\downarrow$       & Enc.[ms]$\downarrow$    & Dec.[ms]$\downarrow$ & FLOPs/pixels$\downarrow$ \\
    \midrule
Factorized & E2E   & $\surd$     & -47.34\%        & \textbf{24.3}           & 35.7                & 42.175K \\
Hyperprior & E2E   & $\surd$     & \underline{-57.58\%}        & 141           & 169                & 45.546K \\
COIN & RM &                      & 25.53\%                  & $1.40\times10^6$        & 3.18                & 29.057K\\
Cool-Chic v4.2 & RM   &      & \textbf{-64.86\%}        & $1.77\times10^5$           & 216                & 1.303K \\
GaussianImage  & RM &            & 36.55\%                  & $6.99\times10^5$        & \underline{0.439}   & - \\
PIC (Our) & E2E RM &  $\surd$    & -12.78\%     & \underline{28.6}        & \textbf{0.418}      & \textbf{1.074K} \\

    \bottomrule

\end{tabular}
    }
\label{intro:concept}
\end{table}

In this paper, we propose a new image compression paradigm, Practical INR Image Codec (PIC), which bridges the gap between these paradigms. PIC directly produces a low-bit-rate neural representation in a single forward pass through an end-to-end trained network, combining fast encoding, ultra-fast decoding, and competitive rate-distortion performance. 
\cref{intro:concept} demonstrates the main differences among three paradigms.

The primary contributions of this work are summarized below:
\begin{itemize}
    \item We propose a novel image coding paradigm PIC that directly generates low-bit-rate neural representation models through neural networks, simultaneously achieving fast encoding, ultra-fast decoding, and comparative RD performance.
    \item We have implemented an optimized decoder that fully translates the low-complexity characteristics of neural representation models into practical high decoding speeds.
    \item Through comprehensive experiments, we validate the performance of our proposed method, demonstrating significant improvements in both RD performance and encoding/decoding speed compared to prior representation-based image coding approaches. Notably, our method surpasses nvJPEG in terms of decoding speed.
\end{itemize}
\section{Related Work}

\subsection{End-to-end Image Compression}
Classic end-to-end image compression extends transform coding paradigm, which use neural network as both analysis transform and synthesis transform \cite{balle2017end}.
An important feature that distinguishes compression models from other models is the integer symbol constraints in entropy coding, which introduces non-differentiable quantization in training \cite{guo2021soft}. Ball{\'e} \etal~\cite{balle2017end} pioneered a method to jointly optimize reconstruction loss and bit-rate constraints by maximizing the Evidence Lower Bound (ELBO) and formalizes the compression model optimization task as
\begin{equation}
    \mathcal{L}_{\phi_g, \theta_g} = \lambda R + D(g_s(Q(g_a(\boldsymbol{x};\phi_g)); \theta_g), \boldsymbol{x}),
\end{equation}
where $g_a(\cdot;\phi_g) $ and $g_s(\cdot;\theta_g) $ are analysis transformation and synthesis transformation respectively. $Q(\cdot)$ is quantization operation. To achieve better task performance, several previous works have also explored many differentiable approximation of quantization \cite{balle2017end, guo2021soft}. $R$ is estimated bit rate. $\lambda$ balances the reconstruction quality and bit-rate.

This fundamental architecture has been extensively developed in subsequent research. One important approach is to explore more efficient network architectures. Ball{\'e} \etal~\cite{balle2018variational} introduced scale hyperpriors to better model latent distribution. More work investigate auto-regressive structure \cite{2018Joint, 2020Channel, 2022ELIC} or Transformer model \cite{zou2022devil, liu2023tcm}. Generative models represent another important category. Mentzer \etal~\cite{mentzer2020high} leveraged GAN architectures for high-fidelity reconstruction. With the rise of diffusion models, more research has begun exploring compression in extremely low bit-rate scenarios \cite{xiadiffpc}.


Although end-to-end methods are limited in practical applications due to computational constraints, their single forward encoding capability offers distinct advantages over INR training-based encoding, and provides valuable inspiration for exploring similar INR encoding schemes.

\subsection{Representation-based Image Compression}

These representation-based methods can be further categorized into several types, with the most prominent paradigm being the direct mapping of positional coordinates to target spaces. For instance, NeRF \cite{2020NeRF} maps ray angles to density and color along the ray, which has significantly impacted the fields of volume rendering and novel view synthesis, bringing widespread attention to INR technology \cite{barron2022mip}. Subsequently, this paradigm has expanded to other signal representation domains, including neural rendering \cite{sztrajman2021neural}, image representation, video representation, and further into image and video compression  \cite{ dupont2021coin, chen2021nerv}.

Another approach involves using learnable parameters or grid as inputs instead of directly utilizing coordinates. Within the NeRF series, Instant-NGP \cite{muller2022instant} is a representative work that employs a multi-resolution spatial hash grid to store these learnable parameters. This same concept can be extended to other neural representation tasks. There are many workss have made remarkable progress in representing and compressing textures \cite{vaidyanathan2023random}, BRDF \cite{dou2024real}, etc. Similarly, the COOL-CHIC \cite{ladune2022cool} and its successors \cite{kim2024c3} have demonstrated impressive performance in image compression. This methodology also finds broad applications in video compression \cite{kwan2023hinerv}.

The 3D Gaussian Splatting (3DGS) \cite{kerbl20233d}  provides a different perspective of representation models. By directly organizing learnable parameters through specific structures, without relying on per-instance overfitted neural networks, 3DGS has demonstrated unique advantages in novel view synthesis tasks. Compared with INR, this representation method often exhibit more interpretable structure, providing unique advantages beyond quantity performance. Similarly, such methods have been successfully applied to tasks including 3D scene representation \cite{lu2024scaffold, huang20242d}, image compression \cite{zhang2024gaussianimage}, and video compression \cite{liuexploration}.

This class of methods has demonstrated unique advantages in compression, such as low decoding complexity, fast decoding speed, and high reconstruction quality. However, these strengths are difficult to achieve simultaneously in a single model. Moreover, the reliance on model training for encoding hinders their practical deployment. In this work, our proposed method explores how to balance multiple metrics to construct a practical encoder.

\subsection{Hypernetwork for INR}

Previous work has explored the concept of hypernetworks~\cite{ha2016hypernetworks, klocek2019hypernetwork}, which dynamically modulate neural network weights during inference. This idea can be traced back to even earlier mechanisms such as Fast Weight Programmers (FWP)~\cite{schlag2017gated}, and several studies have also investigated its theoretical connections to linear Transformers~\cite{schlag2021linear}. Existing research on hypernetworks has predominantly focused on tasks such as continual learning~\cite{von2019continual}, few-shot learning~\cite{sendera2023hypershot} and domain adaptation~\cite{volk2022example}. In the field of INR related research, the implementation of generalizable INR generation itself can be formulated as a hypernetwork model. This differs to some extent from the task of generalizable 3DGS generation~\cite{chen2024mvsplat}: 3DGS is, to a certain degree, an explicit representation model, whose distribution characteristics are closely coupled with the final target information. In contrast, the network parameter space of an INR lacks such a correlation with the signal it encodes, which poses a critical challenge for the design of generalizable INR frameworks. In terms of specific algorithm design, hypernetwork-based INR approaches have been explored in various tasks including content generation~\cite{you2023generative, skorokhodov2021adversarial}, yet their investigation in compression tasks remains insufficient~\cite{catania2023nif}.

This paper addresses this gap by further exploring a generalizable/feedforward INR image encoder based on the hypernetwork mechanism. In particular, targeting the lack of research that simultaneously considers RD performance and coding efficiency, we attempt to design a codec that comprehensively considers performance across all dimensions through this approach, offering new insights into the field of learned image codecs.
\section{Method}
\label{sec:method}
\begin{figure}[tb]
\begin{center}
\includegraphics[width=\linewidth]{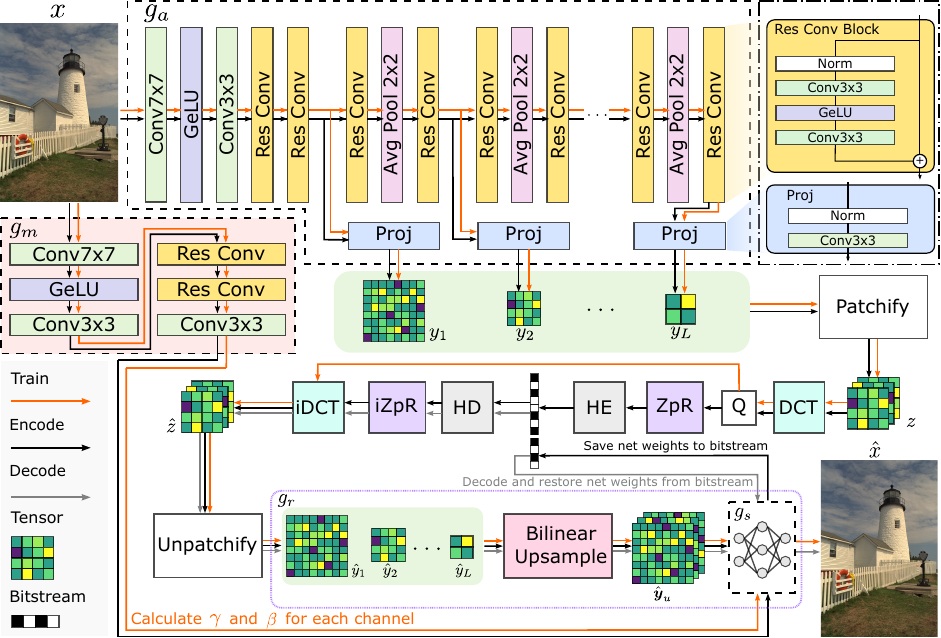}
\end{center}
\caption{The framework of PIC. The overall data flow is presented in an S-shaped in the diagram, following $g_a\rightarrow \hat{\boldsymbol{y}}\rightarrow z \rightarrow \hat{z} \rightarrow g_r \rightarrow g_s$. $g_a$ and $g_m$ are latent encoder and modulation net respectively. $\hat{\boldsymbol{y}} = \{y_1, y_2, \dots y_L\}$ are latents generated by latent encoder $g_a$. 
All latents are divided into patches of size $8\times 8$ and then concatenated together as $z$. Similar to other compression methods, $z$ is quantized after DCT transformation. The ZpR module converts the quantized symbols into a more compact form, and we will discuss this module in detail in the \cref{sec:method:pipeline}. HE and HD are Huffman encoder and decoder. iZpR and iDCT are inverse transformation of DCT and ZpR respectively. 
$g_r$ is part of the decoder and also acts as an image representation model. $g_s$ is synthesis network. $\gamma$ and $\beta$ are mean and standard deviation of each channel of $g_m$ output respectively. 
}
\label{fig:pic_pipeline}
\end{figure}

The primary reason for the slow encoding of representation models lies in the fact that the encoding process itself is the training process. Consequently, methods based on implicit neural representations—as well as similar approaches like Gaussian splatting—can only be applied in scenarios where encoding time is highly insensitive.
A straightforward idea is to directly generate this representation model itself through a neural network. The proposed PIC follows the idea.
\cref{fig:pic_pipeline} demonstrates the overall architecture of our method. 
\cref{sec:method:r2c} shows the details of transforming a representation to compression models. \cref{sec:method:entropy} introduces entropy estimation module. \cref{sec:method:pipeline} describes the full pipeline and implementation details.

\subsection{Repurposing Representation Models for Compression}
\label{sec:method:r2c}
To better introduce the proposed method, we begin with the image representation model $g_r$ in \cref{fig:pic_pipeline}. For a $H\times W$ image,
$\hat{\boldsymbol{y}}$ is a set of pyramid-like multi-resolution latents
\begin{equation}
\hat{\boldsymbol{y}}=\{\hat{y}_i\in\mathbb{R}^{H_i\times W_i}, i = 1, 2, \ldots, L\},
\end{equation}
where $H_i = \frac{H}{2^{L - i}}, W_i = \frac{W}{2^{L - i}}$. $L$ is the number of latents.
To match the resolution of the output image, $\hat{\boldsymbol{y}}$ is upsamlped to $\hat{\boldsymbol{y}}_u\in \mathbb{R}^{C\times H\times W}$ before being fed into the reconstruction network $g_s$.
If we apply $g_r$ in image representation task, image information will be stored in the weights of the network. All parameters, including $\hat{\boldsymbol{y}}$ and weights in $g_s$, are trainable and optimized jointly via gradient descent. 

Obviously, achieving a fast encoding process through training is highly challenging. Therefore, our approach generates all weights of the representation model in a single step. Since the latents inherently lies in a space similar to the image domain, designing a simple yet functional encoder $g_a$ is relatively straightforward, as shown in \cref{fig:pic_pipeline}. 
The primary challenge is generating the network weights of $g_s$. While previous works have explored methods for network weight generation, their performance in compression tasks still leaves room for improvement \cite{chen2024fast}. In this paper, we adopt a simple strategy called channel-wise normalization and modulation. 

We first divide $g_s$ into shared part $\mathrm{MLP}^s$ and instance-dependent part $\mathrm{MLP}^i$.
Suppose $N$ is batch size and $\boldsymbol{f}\in \mathbb{R}^{N\times C\times H\times W}$ is the intermediate feature between  $\mathrm{MLP}^s$ and $\mathrm{MLP}^i$, the instance-normalized feature is
\begin{equation}
    \bar{\boldsymbol{f}} = \frac{\boldsymbol{f} - \mathrm{mean}(\boldsymbol{f})}{\mathrm{std}(\boldsymbol{f})}.
\end{equation}
Then modulate the normalized feature $\bar{\boldsymbol{f}}$
\begin{equation}
    \tilde{\boldsymbol{f}} = \gamma\odot\bar{\boldsymbol{f}} + \beta,
\end{equation}
where $\gamma\in \mathbb{R}^{N\times C}$ and $\beta\in \mathbb{R}^{N\times C}$ are generated by modulation net $g_m$. Fraction and $\odot$  represent element-wise division and multiplication at $C$ dimension respectively. Note $g_m$ will generate corresponding $\gamma$ and $\beta$ for each input $x$.
One notable advantage of this transformation is that both the normalization and modulation parameters can ultimately be fused into the network weights of $\mathrm{MLP}^i$, thereby achieving the goal of generating corresponding network weights for each input. For detailed derivation, please refer to the Supplementary.

\subsection{Entropy Estimation}

\label{sec:method:entropy}


In neural representation models, constrained by the overall model size, it is challenging to incorporate a large entropy estimation network. One solution is to employ a compact neural network to enhance RD performance via auto-regressive methods~\cite{ladune2022cool}. However, the decoding speed of such approaches remains limited by their auto-regressive design. Inspired by Luo \etal~\cite{luo2020rate}, we estimate the distribution of DCT parameters to achieve rate estimation accordingly.


For latents generated by $g_a$
\begin{equation}
\boldsymbol{y}=\{y_i\in\mathbb{R}^{H_i\times W_i}, i = 1, 2, \ldots, L\}, 
\end{equation}
we divide them into patches of size $8\times 8$ , concatenate together as $z\in\mathbb{R}^{n\times 8\times 8}$ and transform to symbols through DCT 
\begin{equation}
    z = \mathrm{Patchify}(y_1, \dots, y_L), 
\end{equation}
\begin{equation}
    s = \mathrm{DCT}(z).
\end{equation}
Let $s_{i,k}$ is the $k$-th DCT component ($k \in \{0, \dots, 63\}$) of $i$-th block, the estimated entropy is
\begin{equation}
   \mathcal{L}_\text{entropy} = - \sum_{i=1}^n\sum_{k=0}^{63}\log_2 p_{\theta_k}(s_{i,k} + u), u\sim \mathrm{Uniform}(-0.5, 0.5).
\end{equation}
We use 64 channels entropy model, which means we model the distribution of DCT components independently. $\theta_k$ is the corresponding parameters of the piecewise linear function for the $k$-th channel in entropy model~\cite{balle2017end,Jean2020CompressAI}.
Similar to other compression framework , the entire pipeline is non-differentiable after quantization of $s$ , so we use a simple uniform noise relaxation \cite{balle2017end} to build an end-to-end trainable pipeline.




\subsection{Full Pipeline and Hardware-affinity Implementation}
\label{sec:method:pipeline}
\begin{figure}[t]
\begin{center}
\includegraphics[width=0.8\linewidth]{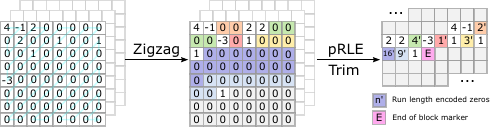}
\end{center}
\caption{The pipeline of ZpR module. ZpR transform quantized DCT coefficients to compact symbols through three steps: zigzag reorder, trim trailing zeros and encode zeros using run length encode (partial RLE).}
\label{fig:zpr}
\end{figure}

JPEG reduces bit-rate through lossless compression techniques like the zigzag transform and run-length encoding (RLE). 
In our PIC, we similarly employ \textbf{Z}igzag reordering and \textbf{p}artial \textbf{R}un-length encoding (ZpR) module to improve RD performance.
As shown in \cref{fig:zpr}, ZpR reduce zeros in zigzag reordered symbols. The main difference between ZpR and RLE in JPEG is we only reduce zeros in symbols.
This strategy is more straightforward to implement while effectively reducing the bit-rate.

Unlike JPEG, our method employs image-specific Huffman coding tables, meaning the storage overhead of the Huffman tables also impacts the final bit-rate performance. To ensure $O(1)$ complexity for decoding one symbol, the coding table includs $2^p$ entries ($p$ is precision) which covers all possible bitstream pattern for decoding one symbol. For small pictures like those in Kodak dataset, the overhead of storing the code tables is non-negligible. Fortunately, the actual number of symbols used in encoding is typically fewer than 128, allowing for further compression of the code tables to reduce this overhead. Benefiting from the prefix code property of Huffman coding, we can apply RLE to compress the original code table, reducing the number of table entries to match the symbol count and significantly decreasing the storage overhead of the original code table.


As illustrated in \cref{fig:pic_pipeline}, we integrated all of the above modules to achieve an trainable codec. The final loss function is
\begin{equation}
    \mathcal{L} =  \lambda \mathcal{L}_\text{entropy} + D(x, \hat{x}),
\end{equation}
where $D$ is distortion metric, $\lambda$ balances the trade-off between reconstruction quality and bit-rate.

Another important topic is how to implement an optimized decoder. JPEG's concise design and long-term optimizations have made it one of the fastest encoders. However, due to its early introduction, JPEG has not fully leveraged the hardware advancements in recent years. Benefiting from recent research on parallel Huffman decoding \cite{weissenberger2018massively} and the emergence of hardware features like Tensor Cores, it has become possible to develop a decoder that surpasses JPEG in performance.

In architecture design, we avoided using relatively large synthesis network like COOL-CHIC-family \cite{ladune2022cool,kim2024c3} to comply with Tensor Core's matrix structure requirements. Each layer in our reconstruction network $g_s$ maintains a width of 16, precisely matching the warp matrix multiplication of TF32 operations in Tensor Cores. This alignment maximizes hardware utilization and accelerates reconstruction speed. It should be noted that, although our current implementation relies on NVIDIA's Tensor Cores, other hardware also supports similar types of functionalities, such as AMD's rocWMMA. We believe such acceleration hardware will eventually be supported by more manufacturers' devices in the future as well.

Furthermore, since the precision of TF32 is lower than standard FP32, a gap would arise if training were conducted with FP32 while inference/decoding relied on Tensor Core-accelerated operations. To address this, we draw inspiration from neural rendering techniques and implement the training-phase code using the differentiable shading language Slang \cite{he2018slang}, which significantly reduces the complexity of integrating inline CUDA code in differentiable framework. This approach ensures consistency between training and inference while maintaining computational efficiency during training.

Qualitatively speaking, our decoder's complexity is actually higher than JPEG's. However, by fully leveraging hardware capabilities, our decoder has surpassed commercial decoders and provided new insights for the design of image codecs.




\section{Experiments}


\label{sec:exp}

 \begin{figure}[t]
    \center
  \subfloat[Rate-distortion performance on Kodak dataset.]{\label{fig:main_results:q}\includegraphics[width = 0.98\textwidth, trim=100 0 100 0,clip]{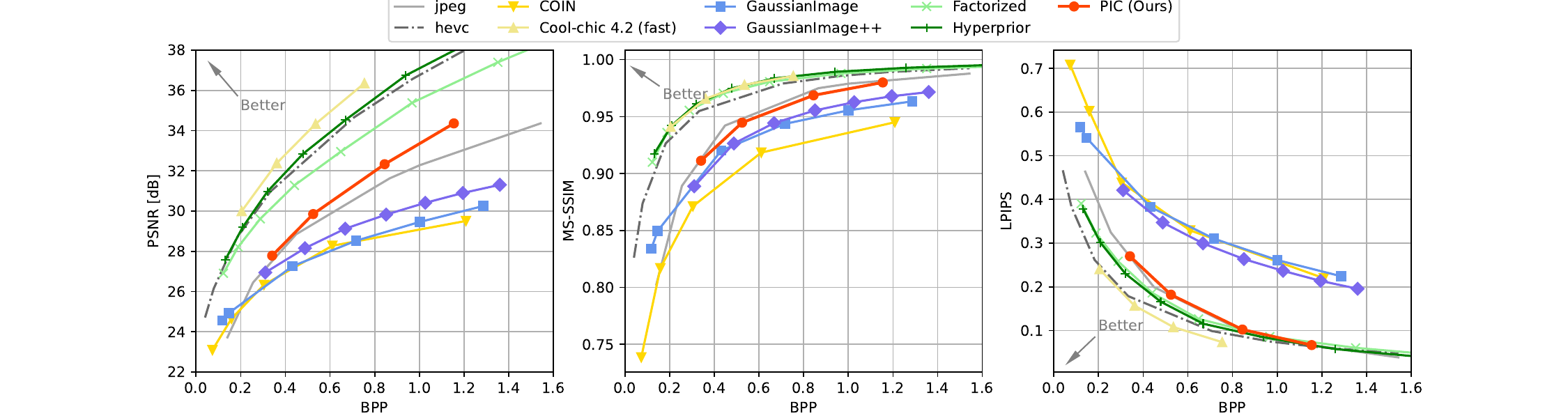}}


  \subfloat[Speed and complexity comparison of all methods on Kodak dataset.]{\label{fig:main_results:s}\includegraphics[width = 0.98\textwidth, trim=100 0 100 0,clip]{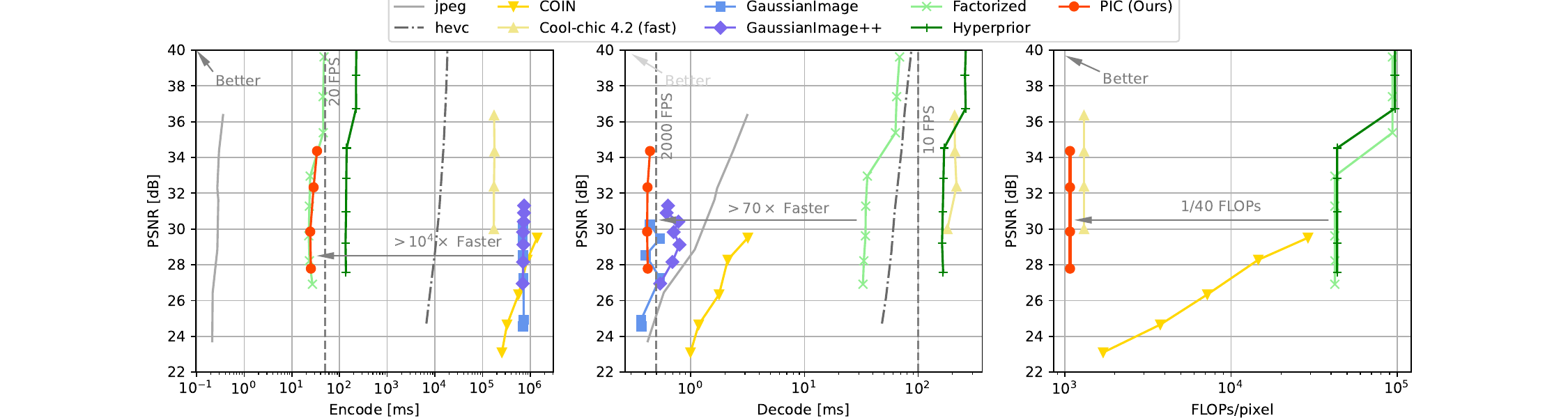}}


  \caption{Results on Kodak dataset. 
  }
    \label{fig:main_results}
\end{figure}
\begin{figure}[t]
    \center
  \subfloat[Rate-distortion performance on CLIC dataset.]{\label{fig:clic_results:q}\includegraphics[width = 0.98\textwidth, trim=100 0 100 0,clip]{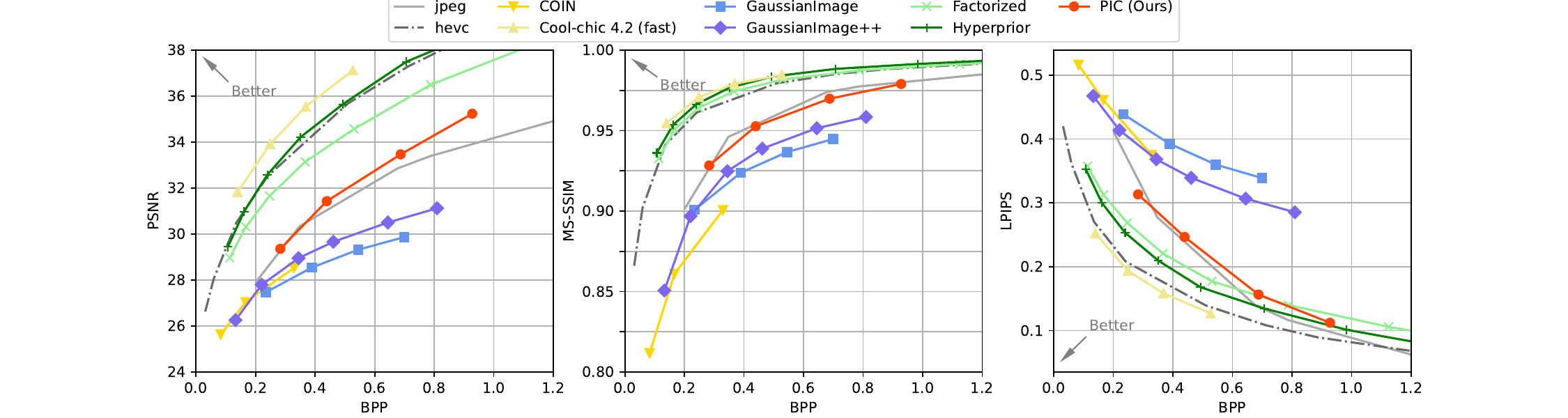}}


  \subfloat[Speed and complexity comparison of all methods on CLIC dataset.]{\label{fig:clic_results:s}\includegraphics[width = 0.98\textwidth, trim=100 0 100 0,clip]{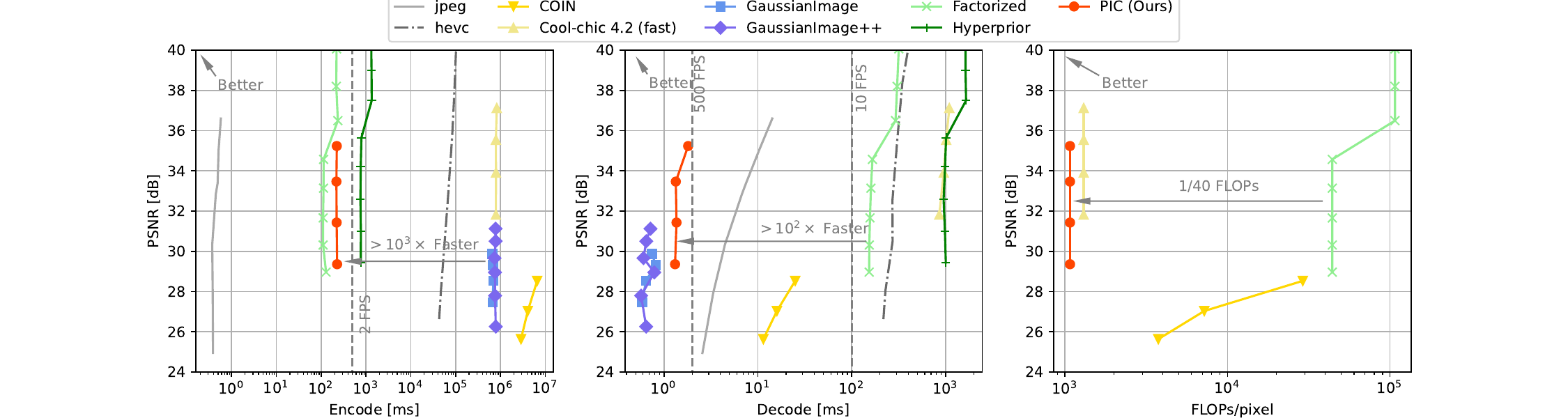}}


  \caption{Results on CLIC dataset.}
    \label{fig:main_results:clic}
\end{figure}

\subsection{Experiment Setup}
\label{sec:exp:setting}
\textbf{Dataset.} Since the proposed PIC is a generalizable method, it is necessary to train our model on a large dataset like end-to-end methods.
We use LSDIR dataset \cite{li2023lsdir} as training set, which includes 84,991 natural images.  During training, these images were randomly
 cropped to a resolution of $512\times 512$. Additionally, Our evaluation is conducted on two popular datasets, Kodak\footnote{http://r0k.us/graphics/kodak} and CLIC\footnote{https://clic.compression.cc/2021/tasks/index.html}. Kodak dataset includes 24 images of size $768\times 512$. The CLIC dataset contains 41 high-resolution natural images.

\textbf{Metrics.} We use the peak signal-to-noise ratio (PSNR) in RGB 4:4:4 as distortion metric, which are the most widely used metric in compression. We
 also report and MS-SSIM \cite{wang2003multiscale} LPIPS \cite{zhang2018unreasonable} in main experiment, which is a popular perceptual metric to
 measure realism. Bit-per-pixel (BPP) is used as coding efficiency metric. To demonstrates the efficiency of PIC, we report encoding, decoding time and complexity as well.

 \textbf{Implementation Details.} Our complete framework is implemented in PyTorch, with the key distinction that the synthesis network used in training is built upon Slang-torch \footnote{https://github.com/shader-slang/slang-torch} to simplify Tensor Core programming within the PyTorch environment. The ZpR module and decoder are CUDA-accelerated, while the Huffman codec is adapted and modified from GPUHD \cite{weissenberger2018massively}. Other auxiliary components such as bitstream I/O operations are implemented in C++. All experiments are performed on a single NVidia RTX 4090D. We jointly optimize the full set of parameters in $g_a$, $g_m$ and $g_s$.$g_s$ is a 4 layers MLP with ReLU activation. The final bitstream  includes header information, Huffman code table, encoded latents, and parameters of $\text{MLP}^i$ in FP32 format. 

 \textbf{Hyper-parameters settings.}
For the proposed PIC, we use total $L=8$ latents. This means that the minimum size of images we can encode is $256\times 256$.
This also suggests that encoding smaller images requires reducing the value of $L$.
During the training, we set batch size to 16 and randomly cropped the input image to a resolution of
$512\times 512$.  This optimization is performed separately for each $\lambda = \{0.01, 0.005, 0.002, 0.001\}$ using Adam optimizer and learning rate of 1e-4. All convolution layers have 16 channels. 
We trained our model for 10 epochs in all experiments. Mean squard error (MSE) is used as distortion metric in training.

 \textbf{Benchmarks.} Our method is benchmarked against competitive representation based methods like COIN \cite{dupont2021coin} GaussianImage \cite{zhang2024gaussianimage} and GaussianImage++\cite{li2026gaussianimage++}. We also compares with Cool-Chic v4.2 (fast)~\cite{ladune2022cool, leguay2023low, kim2024c3} which is the state-of-the-art method in representation based image codec. We reproduce all the results using the original source code. Another baselines include pretrained Factorized model \cite{balle2017end} and Hyperprior \cite{balle2018variational} in CompressAI \cite{Jean2020CompressAI} and nvJPEG\footnote{https://developer.nvidia.com/nvjpeg}, which is a CUDA-accelerated JPEG codec. We also include HEVC in comparison. Note we use MSE as distortion metrics in loss function for PIC in all experiments. For other methods, we follow their original loss function settings, even if they used a loss function different from MSE.

\begin{figure}[t]
  \captionsetup[subfigure]{justification=centering}
  \subfloat[Origin\\ \quad]{\includegraphics[width = 0.166\textwidth, trim=0 0 0 0,clip]{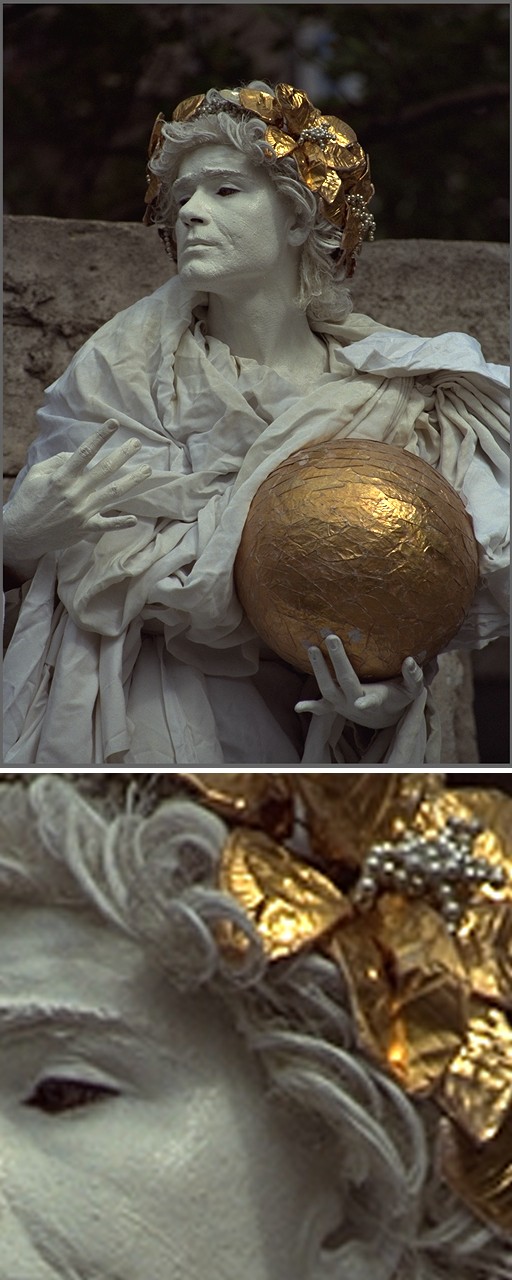}} \hfill 
  \subfloat[JPEG\\34.65 / 1.02]{\includegraphics[width = 0.166\textwidth, trim=0 0 0 0,clip]{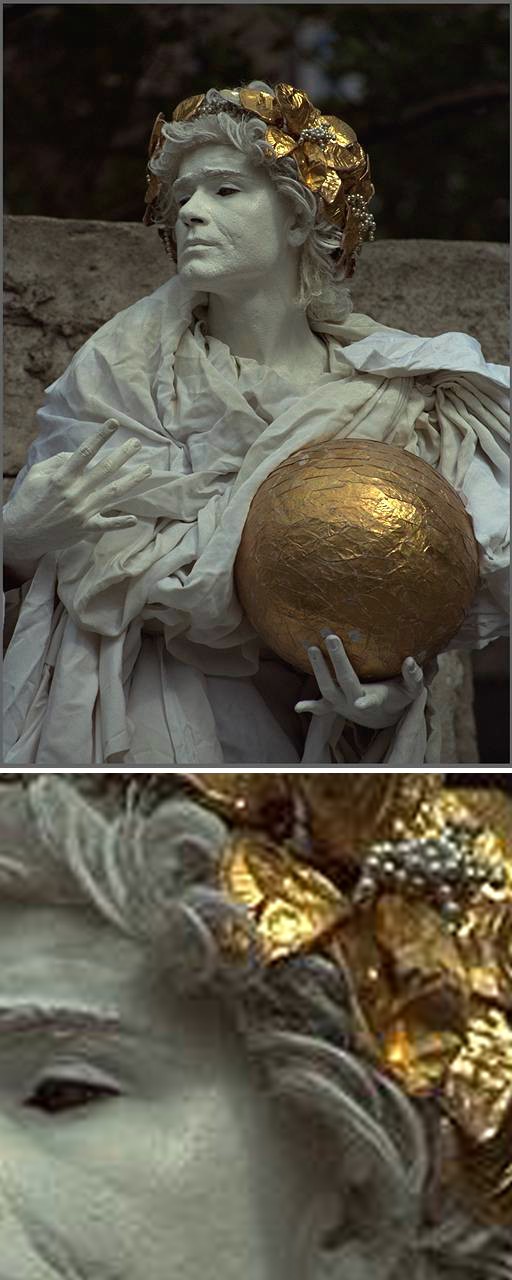}}\hfill 
  \subfloat[Factorized\\36.53 / 0.71]{\includegraphics[width = 0.166\textwidth, trim=0 0 0 0,clip]{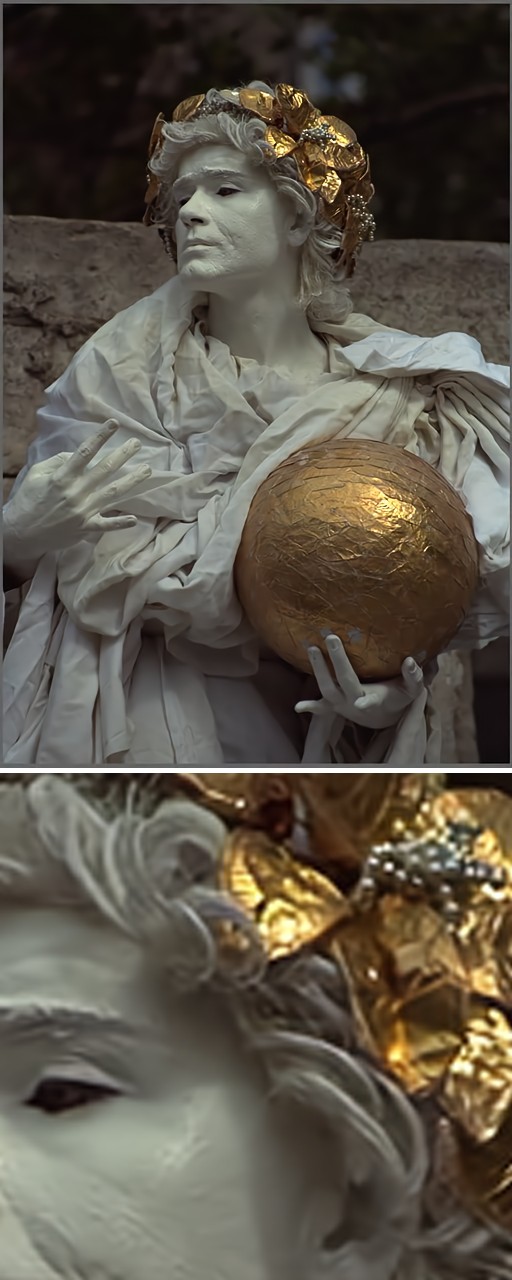}}\hfill 
  \subfloat[COIN\\33.34 / 1.21]{\includegraphics[width = 0.166\textwidth, trim=0 0 0 0,clip]{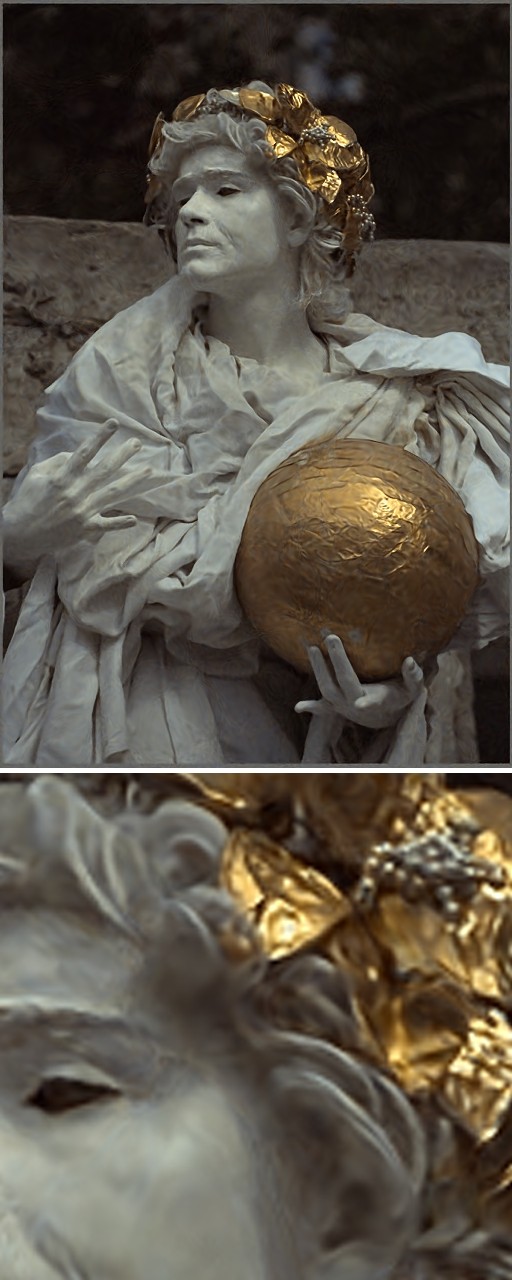}} \hfill 
  \subfloat[GI\\32.41 / 1.29]{\includegraphics[width = 0.166\textwidth, trim=0 0 0 0,clip]{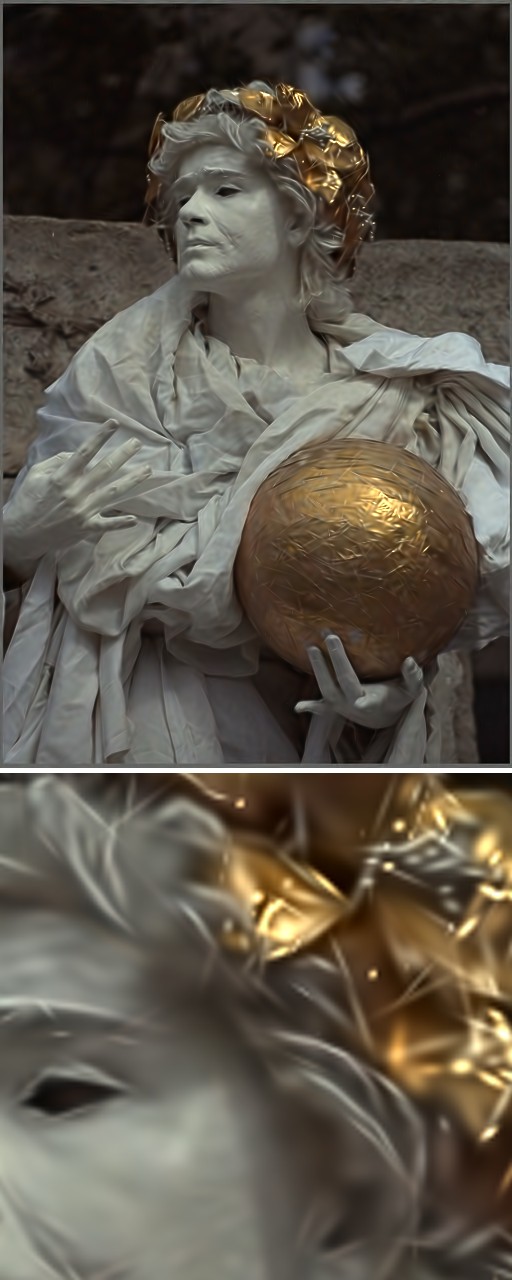}}\hfill 
  \subfloat[PIC\\35.34 / 0.89]{\includegraphics[width = 0.166\textwidth, trim=0 0 0 0,clip]{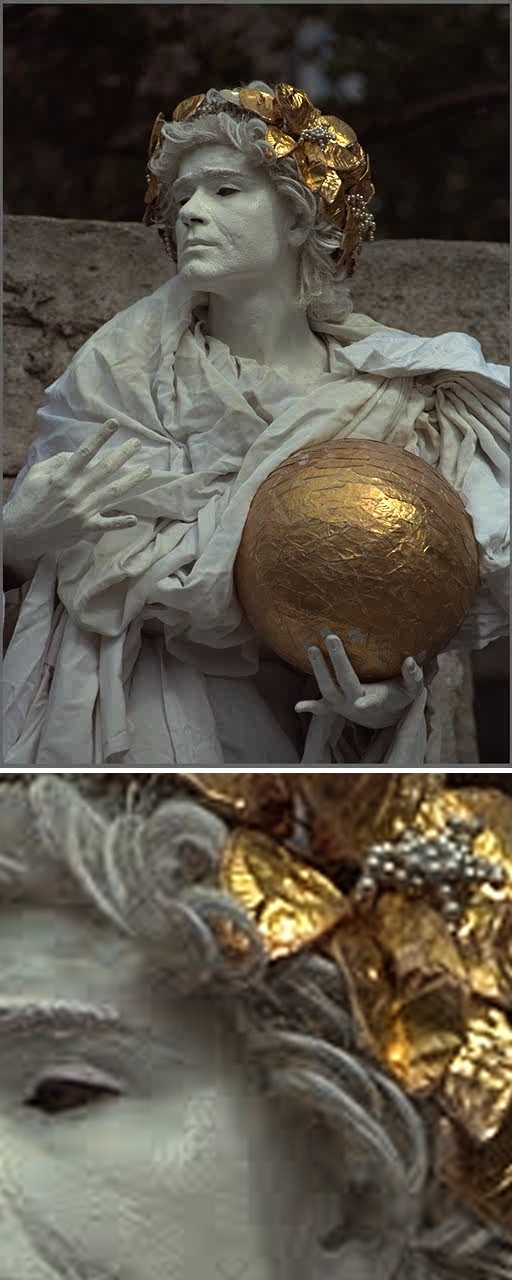}}

  \subfloat[Origin]{\includegraphics[width = 0.166\textwidth, trim=0 0 0 0,clip]{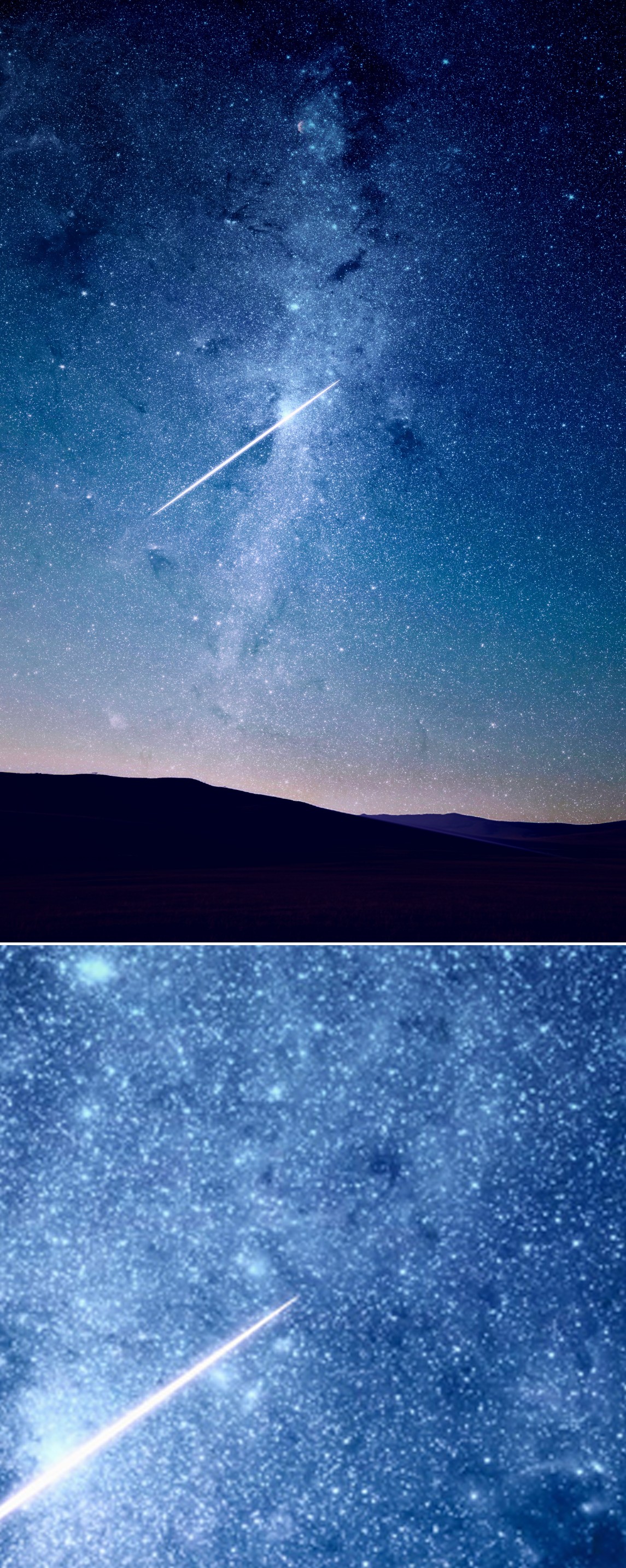}}\hfill 
  \subfloat[JPEG\\31.34 / 1.71]{\includegraphics[width = 0.166\textwidth, trim=0 0 0 0,clip]{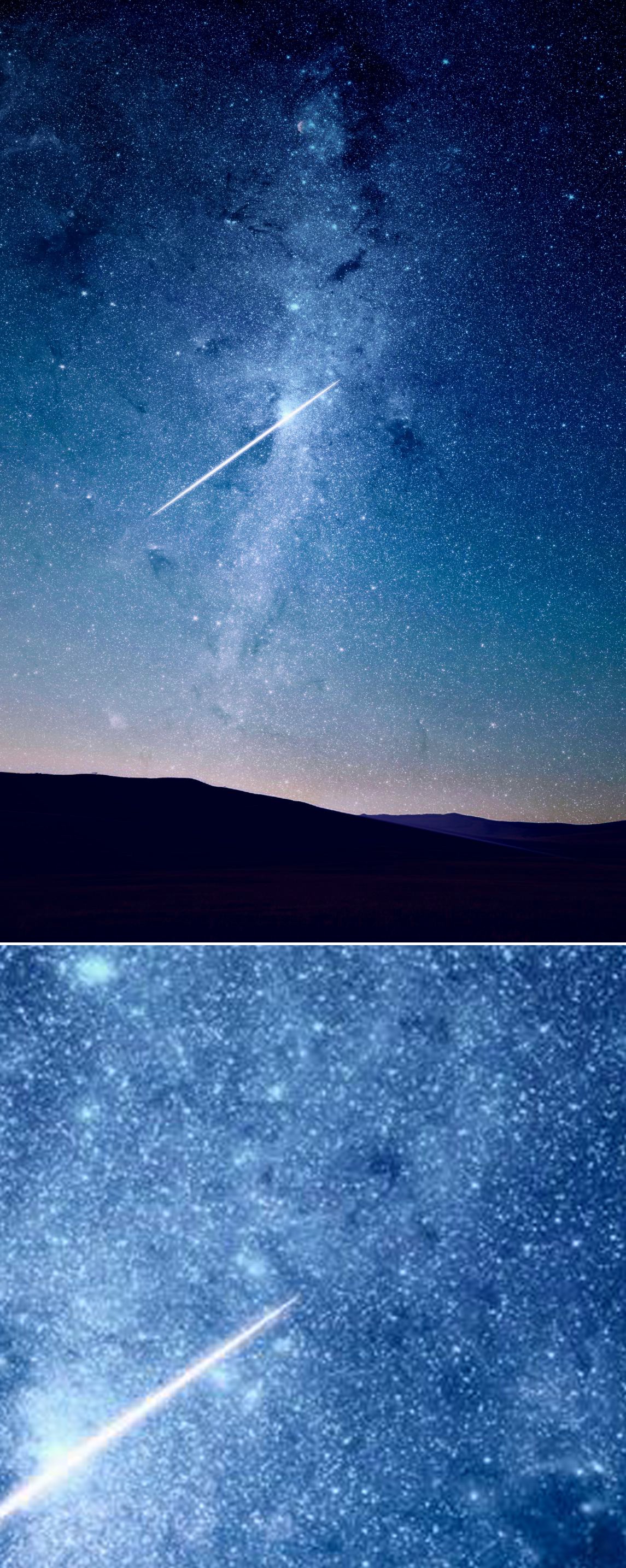}}\hfill 
  \subfloat[Factorized\\32.80 / 1.53]{\includegraphics[width = 0.166\textwidth, trim=0 0 0 0,clip]{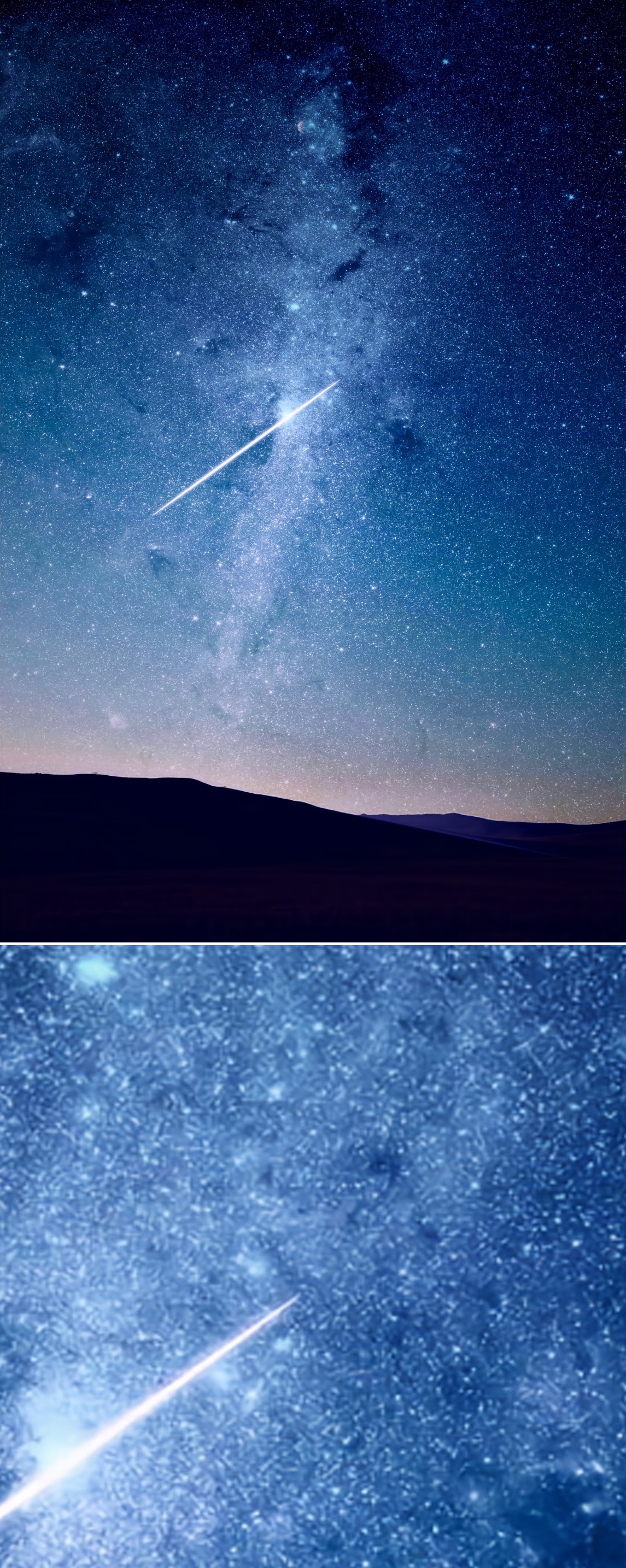}}\hfill 
  \subfloat[COIN\\24.67 / 0.24]{\includegraphics[width = 0.166\textwidth, trim=0 0 0 0,clip]{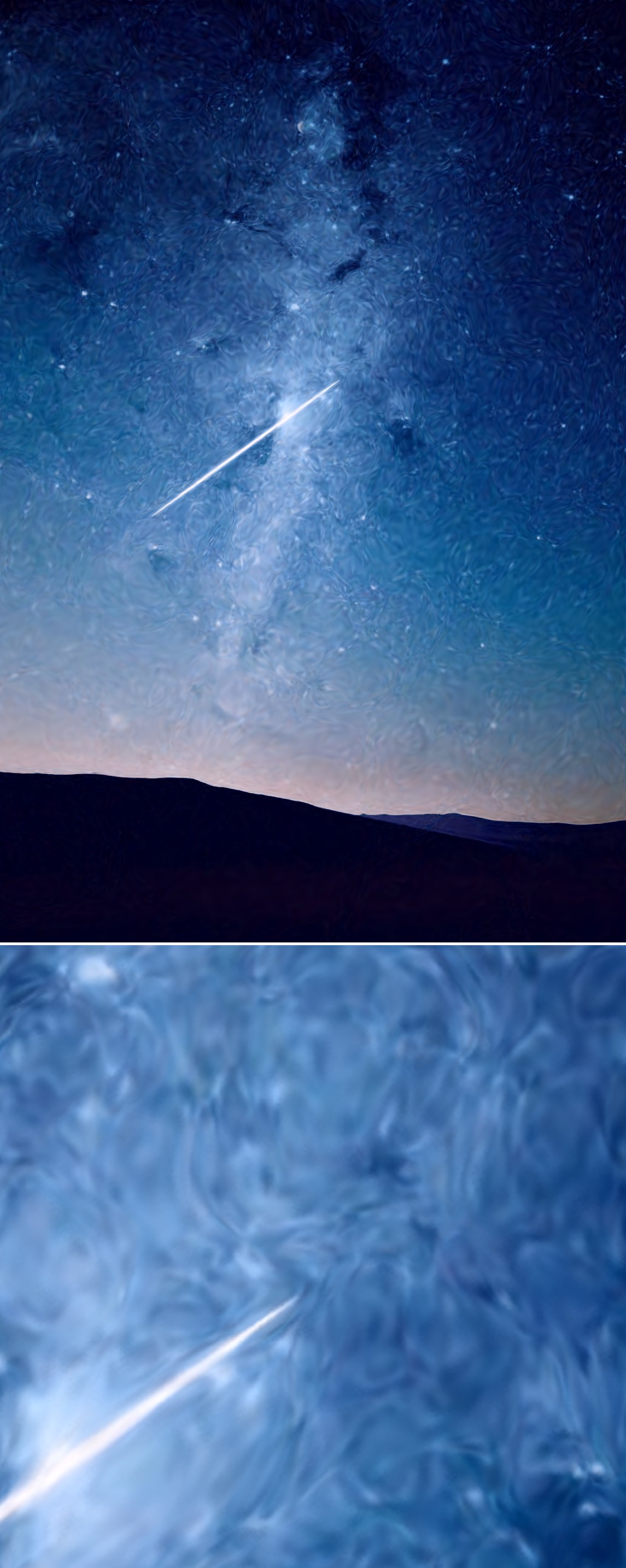}}\hfill 
  \subfloat[GI\\25.27 / 0.51]{\includegraphics[width = 0.166\textwidth, trim=0 0 0 0,clip]{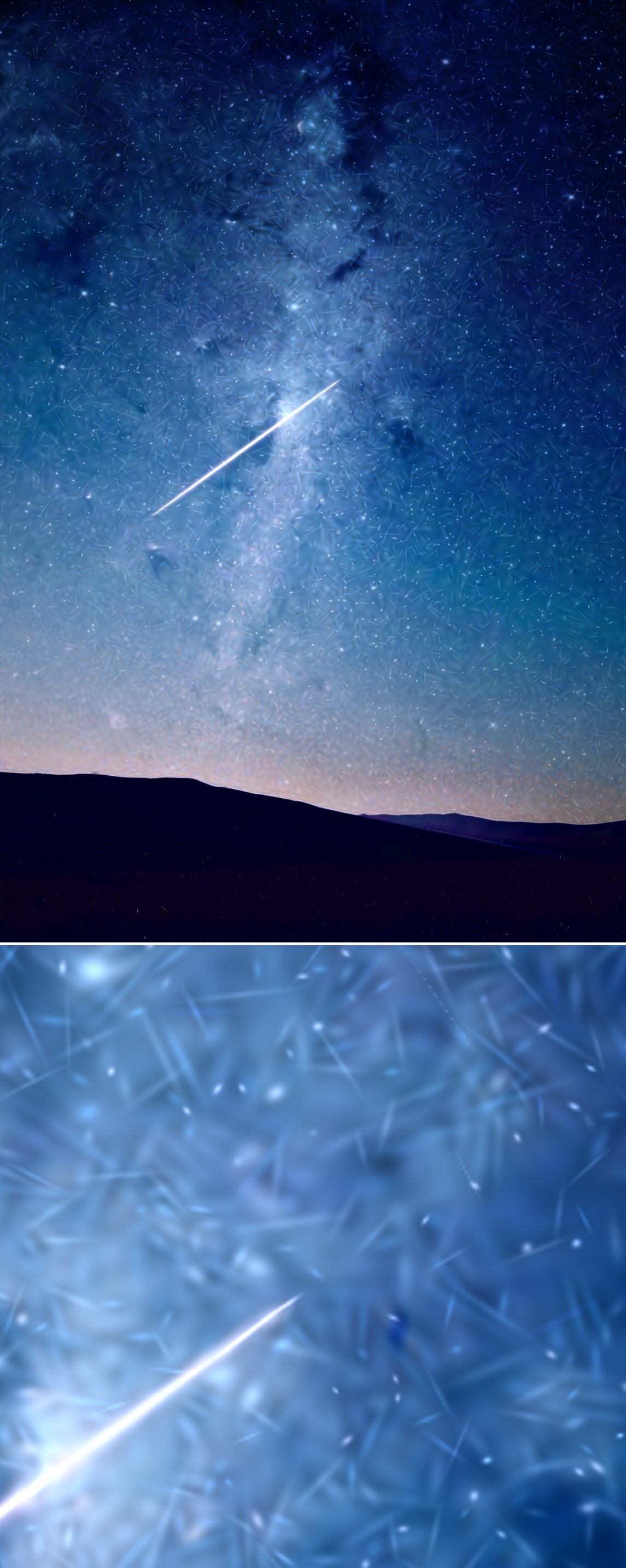}}\hfill 
  \subfloat[PIC\\31.88 / 1.71]{\includegraphics[width = 0.166\textwidth, trim=0 0 0 0,clip]{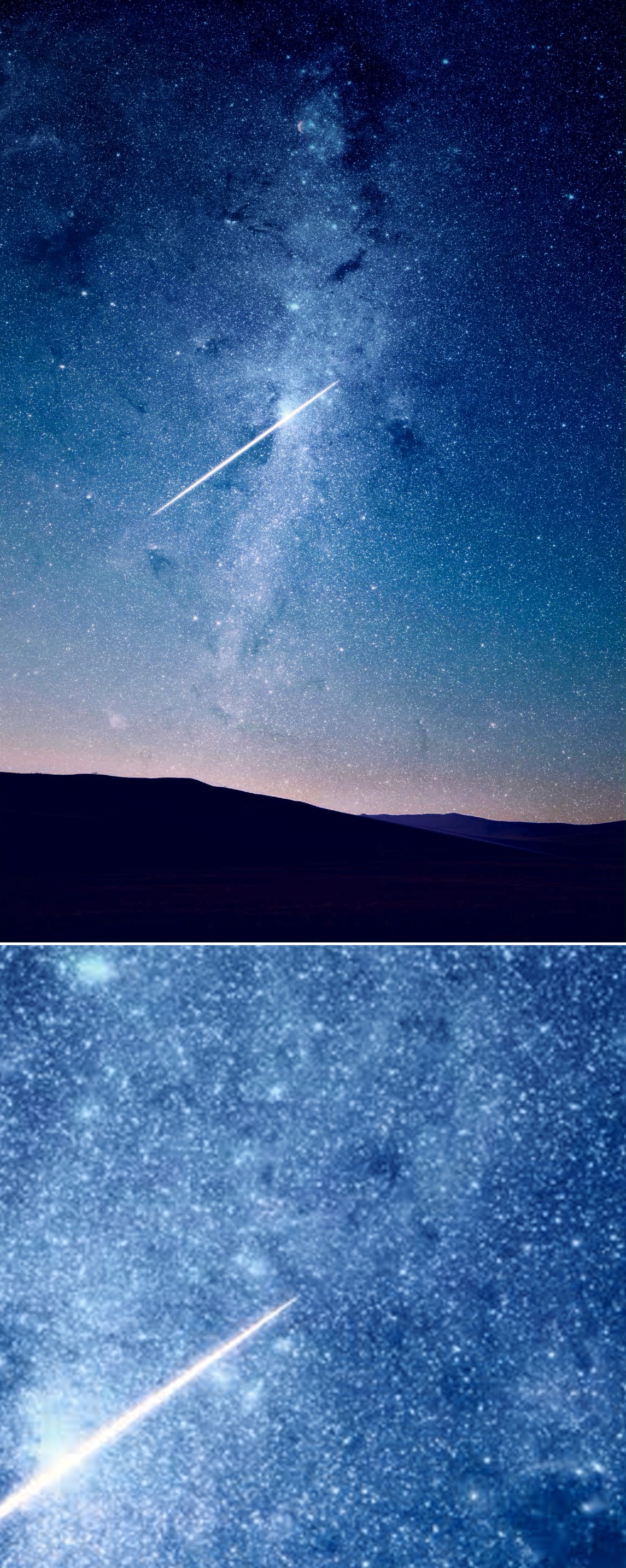}}

  \caption{Visualizations on kodim17.png from Kodak dataset and juskteez-vu-1041.png from CLIC dataset.
    }
    \label{fig:quality_vis}
\end{figure}
 \subsection{Quantity and Quality Results}



\cref{fig:main_results} and \cref{fig:main_results:clic}  shows the main results of different methods on Kodak dataset and CLIC dataset repectively. 
\cref{fig:main_results:q} and \cref{fig:clic_results:q} demonstrate the RD performance of all methods. All results are averaged on same corresponding hyper-parameters setting for each method. 
PIC achieves better PSNR at high bit-rate region and comparable MS-SSIM and LPIPS performance with JPEG. PIC also outperforms other representation-based methods with similar decoding speed in all quality metrics at high bit rate region. 

\cref{fig:main_results:s} and and \cref{fig:clic_results:s} present comprehensive comparisons of encoding/decoding speeds across all methods. PIC achieves comparable encoding speed to Factorized model \cite{balle2017end} and Hyperprior model~\cite{balle2018variational} but significantly faster decoding speed and lower decoding complexity.
Compared with other representation-based approaches~\cite{dupont2021coin,ladune2022cool,zhang2024gaussianimage}, PIC has an acceleration of exceeding three orders of magnitude. In terms of decoding speed, our approach ranks second only to GaussianImage~\cite{zhang2024gaussianimage} while outperforming all other alternatives. 

This not only demonstrates the superiority of our proposed method, but also highlights the unique advantages of grid-based approaches in reconstruction quality. Our experiments further validate the performance advantages of COIN~\cite{dupont2021coin} and GaussianImage~\cite{zhang2024gaussianimage} in the low bit-rate region, while these methods exhibit weaker quality improvement compared to other approaches as the bit-rate increases.

\cref{fig:quality_vis} is the visualization of all methods including JPEG \cite{wallace1992jpeg}, Factorized model \cite{balle2017end}, COIN \cite{dupont2021coin}, GaussianImage (abbreviated as GI) \cite{zhang2024gaussianimage} and proposed PIC. Below each image are the corresponding PSNR/bpp results. On juskteez-vu-1041.png, we select the model setting with best quality in original paper for COIN and  GaussianImage.
Our method demonstrates marginally superior performance compared to JPEG while significantly outperforming COIN and GaussianImage. Our approach also achieves comparable results to JPEG in terms of texture details and artifacts, which can be largely attributed to similar entropy modeling. In contrast, the Factorized model tends to produce overly smoothed outputs. COIN exhibits swirling artifacts, and the GaussiaImage displays Gaussian-like speckled artifacts.

Overall, although there is still a certain gap between our method and state-of-the-art approaches in terms of RD performance, PIC has surpassed the widely used JPEG and is a practically deployable approach. For other learning-based methods, the reality is that current architectures, whether based on end-to-end autoencoders or the overfitted Cool-Chic architecture, are constrained by their inherent limitations, making them difficult to deploy in general scenarios.
Even the most lightweight Factorized model~\cite{balle2017end} is hampered by its high computational overhead. For Cool-chic-like models, their excessively slow encoding and decoding speeds render them impractical for most application. Meanwhile, the proposed PIC exhibits no significant weaknesses across key practical metrics. This offers a new perspective for future research  on learning-based codec in real-world scenarios.







 \subsection{Ablation Study}

 \begin{figure}[t]
  \subfloat[Ablation of synthesis net $g_s$ architecture. All components are necessary for performance.]{\label{fig:syn_ab}\includegraphics[width = 0.49\textwidth, trim=0 0  0 37,clip]{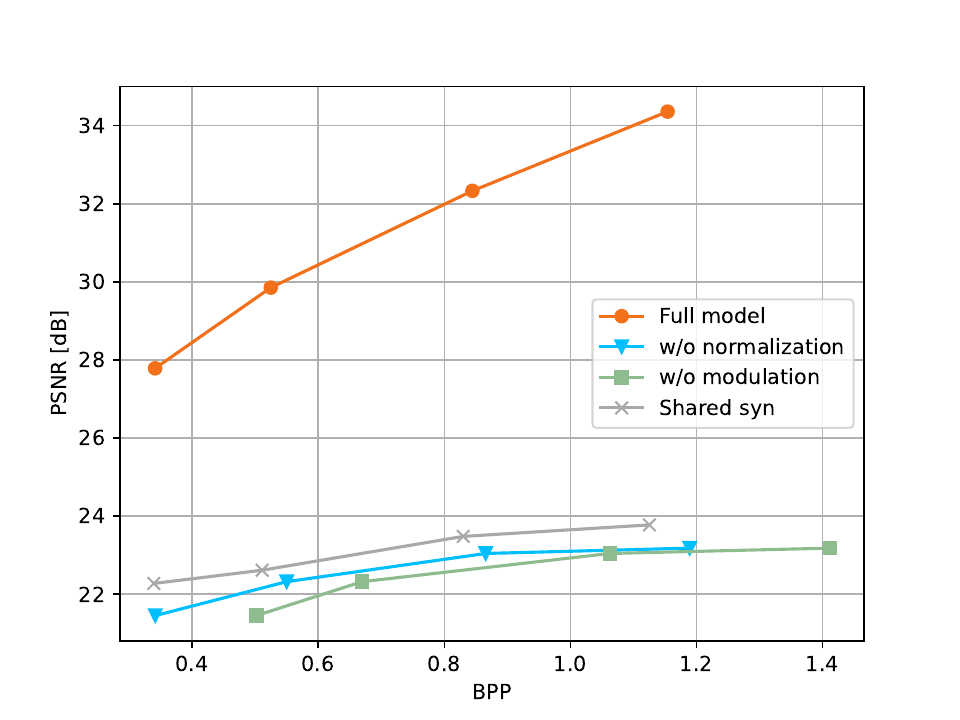}} \hfill 
  \subfloat[Ablation of entropy model. $\rho$ is Pearson correlation coefficient. ]{\label{fig:entropy_ab}\includegraphics[width = 0.49\textwidth, trim=0 0  0 37,clip]{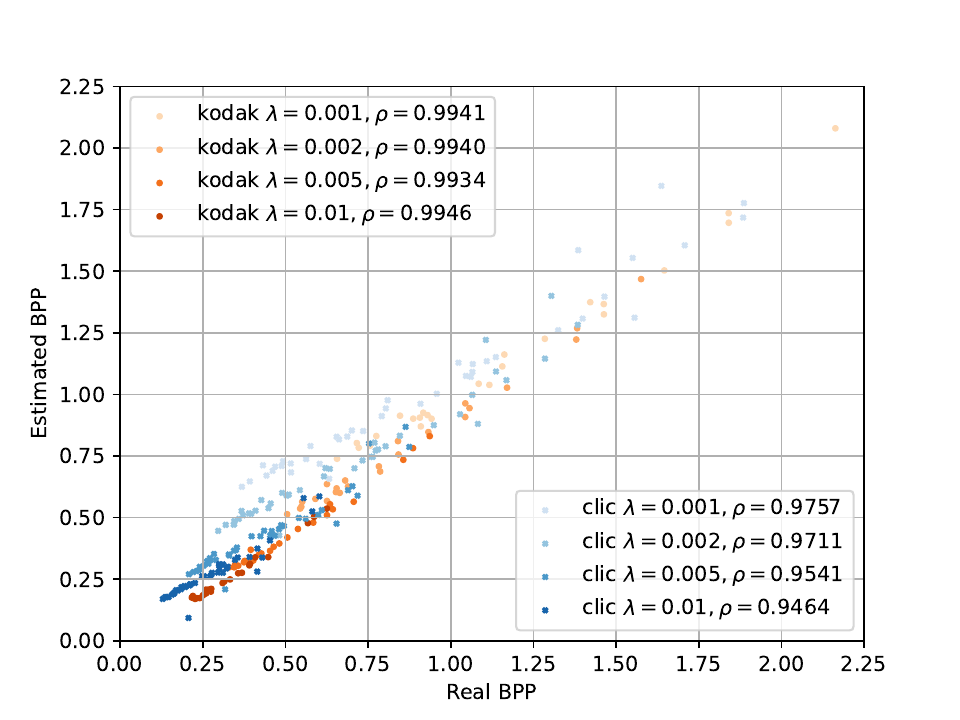}}

  \caption{Ablation results.}
\end{figure}

\cref{fig:syn_ab} presents the architectural ablation results of the synthesis network $g_s$. ``Shared syn'' denotes using the same synthesis net $g_s$ for all images, in which case our method degenerates into an end-to-end model. Evidently, the model performance under this configuration proves significantly inferior due to the limited capacity of decoder network. ``W/o normalization'' and ``w/o modulation'' respectively indicate the exclusion of normalization and modulation in $g_s$. Unsurprisingly, such configurations significantly degrade model performance. Only when incorporating all modules does our method achieve the desired performance.

Due to the inherent approximation nature of entropy networks, discrepancies with actual performance are unavoidable. Furthermore, post-processing in the ZpR module may potentially amplify these deviations. \cref{fig:entropy_ab} illustrates the differences between the actual bit-rate and the estimated bit-rate. Although some discrepancies exist, overall consistency has been maintained. This finding aligns with prior studies \cite{luo2020rate}. Developing more accurate entropy estimation models remains an important direction for future improvements.

 \begin{figure}[t]
    \center

  \subfloat[Decoding speed for different width of synthesis network]{\includegraphics[width = 0.465\textwidth]{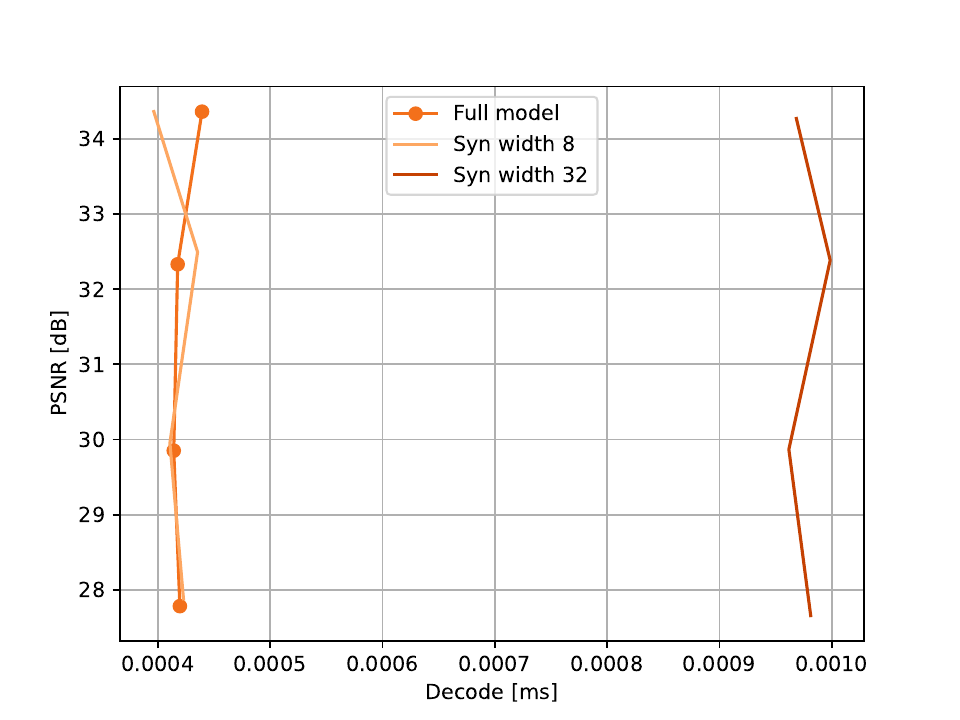}\label{fig:ablation:speed}}
  \subfloat[Detailed breakdown of decoding time]{\includegraphics[width = 0.53\textwidth, trim=20 30 130 0,clip]{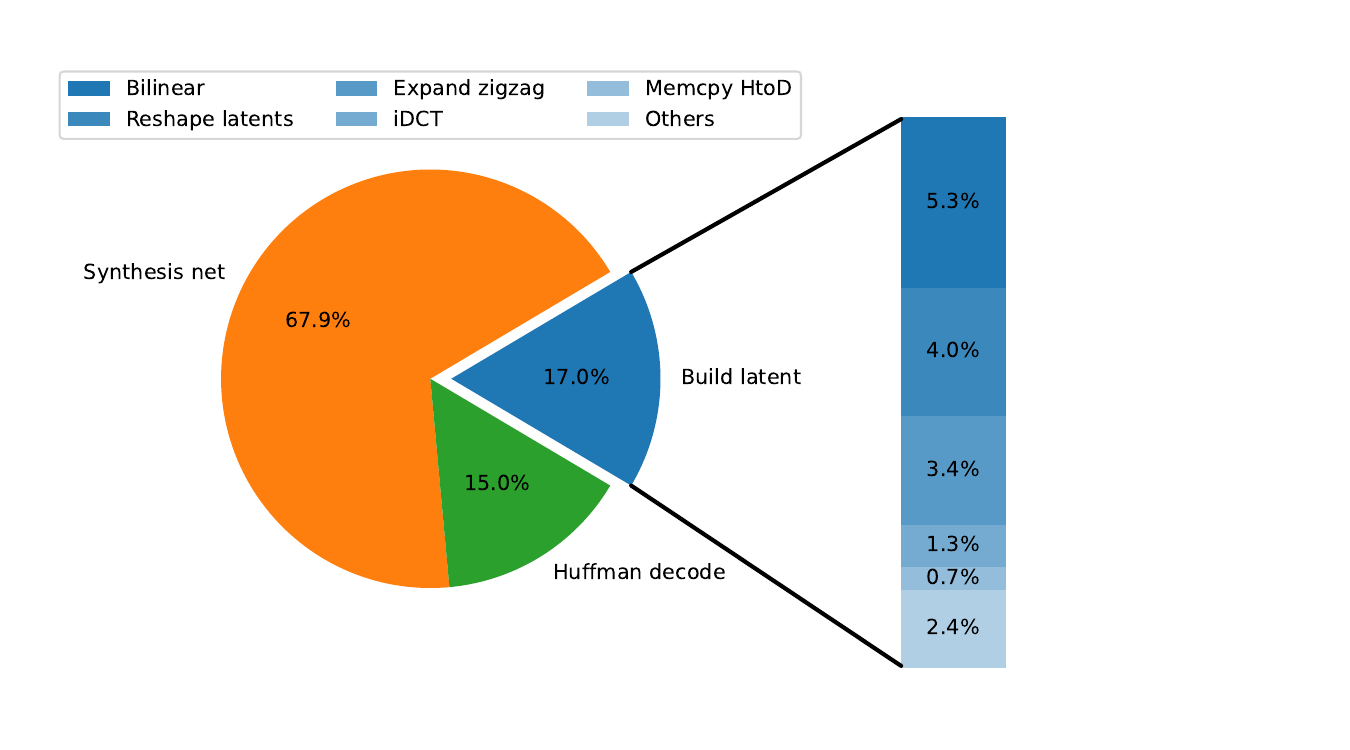}\label{fig:ablation:com}}

  \caption{Ablation results of decoding speed.
    }
    \label{fig:speed_ab}
\end{figure}

\cref{fig:ablation:speed} is the ablation results of decoding speed for different width of synthesis network. Since the minimum matrix size supported by Tensor Cores for the TF32 data type is currently 16, a $16\times 16$ matrix padded with zeros is used when the width is 8. Under this configuration, there is no significant difference in decoding speed between a width of 8 and a width of 16.  Overall, the configuration with a width of 16 is the optimal choice, taking into account decoding speed, RD performance, and hardware compatibility. \cref{fig:ablation:com} is detailed breakdown of decoding time including host-to-device I/O time.

We further analyze the latent representation structure to better understand the model's performance. 
The detailed design of the pRLE module and  the structural design of the synthesis network are also investigated. The visulization of latents and more results are included in the Supplementary. 

\section{Conclusion}
\label{sec:dis}



In this work, we introduced PIC, an end-to-end INR image coding framework that generates all network parameters in a single forward pass, leading to an encoder substantially faster than prior representation-based approaches. We further designed a highly optimized decoder that outperforms JPEG in speed while delivering comparable rate-distortion performance. Together, these advances establish a new balance between efficiency and quality, setting a milestone for INR-based compression. While there is still headroom for improving RD performance, our method opens a novel paradigm for practical learning-based image codecs and provides a foundation for extending INR-based compression to other modalities.

\section*{Acknowledgements}
This work is supported in part by the National Science and Technology Major Project under grant 2025ZD1601300, National Natural Science Foundation of China under grant 62301189, 62576122,62571298, Guangdong Basic and Applied Basic Research Foundation under grant 2026A1515011139.

%
%
\bibliographystyle{splncs04}
\bibliography{main}

@String(AAAI  = {AAAI})

@String(TOG   = {ACM Trans. Graph.})

@String(TOG   = {ACM TOG})

@article{chen2021nerv,
  title={Nerv: Neural representations for videos},
  author={Chen, Hao and He, Bo and Wang, Hanyu and Ren, Yixuan and Lim, Ser Nam and Shrivastava, Abhinav},
  journal={Advances in Neural Information Processing Systems},
  volume={34},
  pages={21557--21568},
  year={2021}
}

@article{kerbl20233d,
  title={3D Gaussian Splatting for Real-Time Radiance Field Rendering.},
  author={Kerbl, Bernhard and Kopanas, Georgios and Leimk{\"u}hler, Thomas and Drettakis, George},
  journal={ACM Trans. Graph.},
  volume={42},
  number={4},
  pages={139--1},
  year={2023}
}

@inproceedings{2020NeRF,
  title={NeRF: Representing Scenes as Neural Radiance Fields for View Synthesis},
  author={ Mildenhall, Ben  and  Srinivasan, Pratul P.  and  Tancik, Matthew  and  Barron, Jonathan T.  and  Ramamoorthi, Ravi  and  Ng, Ren },
  year={2020},
}

@article{dupont2021coin,
  title={Coin: Compression with implicit neural representations},
  author={Dupont, Emilien and Goli{\'n}ski, Adam and Alizadeh, Milad and Teh, Yee Whye and Doucet, Arnaud},
  journal={arXiv preprint arXiv:2103.03123},
  year={2021}
}

@article{dupont2022coin++,
  title={Coin++: Data agnostic neural compression},
  author={Dupont, Emilien and Loya, Hrushikesh and Alizadeh, Milad and Golinski, Adam and Teh, Yee Whye and Doucet, Arnaud},
  journal={arXiv preprint arXiv:2201.12904},
  volume={1},
  number={2},
  pages={4},
  year={2022}
}

@article{zhang2024gaussianimage,
  title={GaussianImage: 1000 FPS Image Representation and Compression by 2D Gaussian Splatting},
  author={Zhang, Xinjie and Ge, Xingtong and Xu, Tongda and He, Dailan and Wang, Yan and Qin, Hongwei and Lu, Guo and Geng, Jing and Zhang, Jun},
  journal={arXiv preprint arXiv:2403.08551},
  year={2024}
}

@inproceedings{lu2024scaffold,
  title={Scaffold-gs: Structured 3d gaussians for view-adaptive rendering},
  author={Lu, Tao and Yu, Mulin and Xu, Linning and Xiangli, Yuanbo and Wang, Limin and Lin, Dahua and Dai, Bo},
  booktitle={Proceedings of the IEEE/CVF Conference on Computer Vision and Pattern Recognition},
  pages={20654--20664},
  year={2024}
}

@inproceedings{ladune2022cool,
  title={Cool-chic: Coordinate-based low complexity hierarchical image codec},
  author={Ladune, Th{\'e}o and Philippe, Pierrick and Henry, F{\'e}lix and Clare, Gordon and Leguay, Thomas},
  booktitle={Proceedings of the IEEE/CVF International Conference on Computer Vision},
  pages={13515--13522},
  year={2023}
}

@inproceedings{wang2003multiscale,
  title={Multiscale structural similarity for image quality assessment},
  author={Wang, Zhou and Simoncelli, Eero P and Bovik, Alan C},
  booktitle={The Thrity-Seventh Asilomar Conference on Signals, Systems \& Computers, 2003},
  volume={2},
  pages={1398--1402},
  year={2003},
  organization={Ieee}
}

@inproceedings{zhang2018unreasonable,
  title={The unreasonable effectiveness of deep features as a perceptual metric},
  author={Zhang, Richard and Isola, Phillip and Efros, Alexei A and Shechtman, Eli and Wang, Oliver},
  booktitle={Proceedings of the IEEE conference on computer vision and pattern recognition},
  pages={586--595},
  year={2018}
}

@inproceedings{liuexploration,
  title={An Exploration with Entropy Constrained 3D Gaussians for 2D Video Compression},
  author={Liu, Xiang and Chen, Bin and Liu, Zimo and Wang, Yaowei and Xia, Shu-Tao},
  booktitle={The Thirteenth International Conference on Learning Representations},
    year={2025}
}

@inproceedings{balle2018variational,
  title={Variational image compression with a scale hyperprior},
  author={Ball{\'e}, Johannes and Minnen, David and Singh, Saurabh and Hwang, Sung Jin and Johnston, Nick},
  booktitle={International Conference on Learning Representations},
  year={2018}
}

@inproceedings{balle2017end,
  title={End-to-end Optimized Image Compression},
  author={Ball{\'e}, Johannes and Laparra, Valero and Simoncelli, Eero P},
  booktitle={International Conference on Learning Representations},
  year={2017}
}

@article{wallace1992jpeg,
  title={The JPEG still picture compression standard},
  author={Wallace, Gregory K},
  journal={IEEE transactions on consumer electronics},
  volume={38},
  number={1},
  pages={xviii--xxxiv},
  year={1992},
  publisher={IEEE}
}

@article{skodras2001jpeg,
  title={The JPEG 2000 still image compression standard},
  author={Skodras, Athanassios and Christopoulos, Charilaos and Ebrahimi, Touradj},
  journal={IEEE Signal processing magazine},
  volume={18},
  number={5},
  pages={36--58},
  year={2001},
  publisher={IEEE}
}

@online{bpg,
  author = {Fabrice Bellard},
  title = {BPG Image format},
  year = 2018,
  url = {https://bellard.org/bpg/},
  urldate = {2023-09-26},
  access={2025-10-10}
  }

@article{mentzer2020high,
  title={High-fidelity generative image compression},
  author={Mentzer, Fabian and Toderici, George D and Tschannen, Michael and Agustsson, Eirikur},
  journal={Advances in neural information processing systems},
  volume={33},
  pages={11913--11924},
  year={2020}
}

@inproceedings{xiadiffpc,
  title={DiffPC: Diffusion-based High Perceptual Fidelity Image Compression with Semantic Refinement},
  author={Xia, Yichong and Zhou, Yimin and Wang, Jinpeng and An, Baoyi and Wang, Haoqian and Wang, Yaowei and Chen, Bin},
  booktitle={The Thirteenth International Conference on Learning Representations},
  year={2025}
}

@inproceedings{jiang2023mlic,
  title={Mlic: Multi-reference entropy model for learned image compression},
  author={Jiang, Wei and Yang, Jiayu and Zhai, Yongqi and Ning, Peirong and Gao, Feng and Wang, Ronggang},
  booktitle={Proceedings of the 31st ACM International Conference on Multimedia},
  pages={7618--7627},
  year={2023}
}

@inproceedings{he2022elic,
  title={Elic: Efficient learned image compression with unevenly grouped space-channel contextual adaptive coding},
  author={He, Dailan and Yang, Ziming and Peng, Weikun and Ma, Rui and Qin, Hongwei and Wang, Yan},
  booktitle={Proceedings of the IEEE/CVF Conference on Computer Vision and Pattern Recognition},
  pages={5718--5727},
  year={2022}
}

@article{liu2024efficient,
  title={An Efficient Implicit Neural Representation Image Codec Based on Mixed Autoregressive Model for Low-Complexity Decoding},
  author={Liu, Xiang and Chen, Jiahong and Chen, Bin and Liu, Zimo and An, Baoyi and Xia, Shu-Tao and Wang, Zhi},
  journal={arXiv preprint arXiv:2401.12587},
  year={2024}
}

@inproceedings{kim2024c3,
  title={C3: High-performance and low-complexity neural compression from a single image or video},
  author={Kim, Hyunjik and Bauer, Matthias and Theis, Lucas and Schwarz, Jonathan Richard and Dupont, Emilien},
  booktitle={Proceedings of the IEEE/CVF Conference on Computer Vision and Pattern Recognition},
  pages={9347--9358},
  year={2024}
}

@inproceedings{guo2021soft,
  title={Soft then hard: Rethinking the quantization in neural image compression},
  author={Guo, Zongyu and Zhang, Zhizheng and Feng, Runsen and Chen, Zhibo},
  booktitle={International Conference on Machine Learning},
  pages={3920--3929},
  year={2021},
  organization={PMLR}
}

@String{Computing = "Computing" }

@String{Computer = "{IEEE} Computer" }

@String{Springer = "Springer-Verlag" }

@ArtifactSoftware{R,
    title = {R: A Language and Environment for Statistical Computing},
    author = {{R Core Team}},
    organization = {R Foundation for Statistical Computing},
    address = {Vienna, Austria},
    year = {2019},
    url = {https://www.R-project.org/},
}

@article{2018Joint,
  title={Joint Autoregressive and Hierarchical Priors for Learned Image Compression},
  author={ Minnen, David  and Johannes Ballé and  Toderici, George },
  year={2018},
}

@misc{2020Channel,
  title={Channel-wise Autoregressive Entropy Models for Learned Image Compression},
  author={ Minnen, David  and  Singh, Saurabh },
  publisher={IEEE},
  year={2020},
}

@article{2022ELIC,
  title={ELIC: Efficient Learned Image Compression with Unevenly Grouped Space-Channel Contextual Adaptive Coding},
  author={ He, Dailan  and  Yang, Ziming  and  Peng, Weikun  and  Ma, Rui  and  Qin, Hongwei  and  Wang, Yan },
  journal={arXiv e-prints},
  year={2022},
}

@article{muller2022instant,
  title={Instant neural graphics primitives with a multiresolution hash encoding},
  author={M{\"u}ller, Thomas and Evans, Alex and Schied, Christoph and Keller, Alexander},
  journal={ACM Transactions on Graphics (ToG)},
  volume={41},
  number={4},
  pages={1--15},
  year={2022},
  publisher={ACM New York, NY, USA}
}

@article{leguay2023low,
  title={Low-complexity Overfitted Neural Image Codec},
  author={Leguay, Thomas and Ladune, Th{\'e}o and Philippe, Pierrick and Clare, Gordon and Henry, F{\'e}lix},
  journal={arXiv preprint arXiv:2307.12706},
  year={2023}
}

@article{vaidyanathan2023random,
  title={Random-Access Neural Compression of Material Textures},
  author={Vaidyanathan, Karthik and Salvi, Marco and Wronski, Bartlomiej and Akenine-M{\"o}ller, Tomas and Ebelin, Pontus and Lefohn, Aaron},
  journal={ACM Transactions on Graphics},
  year={2023},
  publisher={Association for Computing Machinery (ACM)}
}

@article{Jean2020CompressAI,
  title={CompressAI: a PyTorch library and evaluation platform for end-to-end compression research},
  author={Jean Bégaint and Fabien Racapé and  Feltman, Simon  and  Pushparaja, Akshay },
  year={2020},
}

@inproceedings{zou2022devil,
  title={The devil is in the details: Window-based attention for image compression},
  author={Zou, Renjie and Song, Chunfeng and Zhang, Zhaoxiang},
  booktitle={Proceedings of the IEEE/CVF conference on computer vision and pattern recognition},
  pages={17492--17501},
  year={2022}
}

@inproceedings{liu2023tcm,
  author = {Liu, Jinming and Sun, Heming and Katto, Jiro},
  title = {Learned Image Compression with Mixed Transformer-CNN Architectures},
  booktitle = {Proceedings of the IEEE/CVF Conference on Computer Vision and Pattern Recognition},
  pages={1--10},
  year = {2023}
}

@inproceedings{sztrajman2021neural,
  title={Neural BRDF representation and importance sampling},
  author={Sztrajman, Alejandro and Rainer, Gilles and Ritschel, Tobias and Weyrich, Tim},
  booktitle={Computer Graphics Forum},
  volume={40},
  number={6},
  pages={332--346},
  year={2021},
  organization={Wiley Online Library}
}

@inproceedings{dou2024real,
  title={Real-Time Neural BRDF with Spherically Distributed Primitives},
  author={Dou, Yishun and Zheng, Zhong and Jin, Qiaoqiao and Ni, Bingbing and Chen, Yugang and Ke, Junxiang},
  booktitle={Proceedings of the IEEE/CVF Conference on Computer Vision and Pattern Recognition},
  pages={4337--4346},
  year={2024}
}

@article{kwan2023hinerv,
  title={Hinerv: Video compression with hierarchical encoding-based neural representation},
  author={Kwan, Ho Man and Gao, Ge and Zhang, Fan and Gower, Andrew and Bull, David},
  journal={Advances in Neural Information Processing Systems},
  volume={36},
  pages={72692--72704},
  year={2023}
}

@inproceedings{huang20242d,
  title={2d gaussian splatting for geometrically accurate radiance fields},
  author={Huang, Binbin and Yu, Zehao and Chen, Anpei and Geiger, Andreas and Gao, Shenghua},
  booktitle={ACM SIGGRAPH 2024 conference papers},
  pages={1--11},
  year={2024}
}

@inproceedings{chen2024fast,
  title={Fast Encoding and Decoding for Implicit Video Representation},
  author={Chen, Hao and Xie, Saining and Lim, Ser-Nam and Shrivastava, Abhinav},
  booktitle={European Conference on Computer Vision},
  pages={402--418},
  year={2024},
  organization={Springer}
}

@article{luo2020rate,
  title={The rate-distortion-accuracy tradeoff: Jpeg case study},
  author={Luo, Xiyang and Talebi, Hossein and Yang, Feng and Elad, Michael and Milanfar, Peyman},
  journal={arXiv preprint arXiv:2008.00605},
  year={2020}
}

@inproceedings{weissenberger2018massively,
  title={Massively parallel Huffman decoding on GPUs},
  author={Wei{\ss}enberger, Andr{\'e} and Schmidt, Bertil},
  booktitle={Proceedings of the 47th International Conference on Parallel Processing},
  pages={1--10},
  year={2018}
}

@article{he2018slang,
  title={Slang: language mechanisms for extensible real-time shading systems},
  author={He, Yong and Fatahalian, Kayvon and Foley, Theresa},
  journal={ACM Transactions on Graphics (TOG)},
  volume={37},
  number={4},
  pages={1--13},
  year={2018},
  publisher={ACM New York, NY, USA}
}

@inproceedings{li2023lsdir,
  title={Lsdir: A large scale dataset for image restoration},
  author={Li, Yawei and Zhang, Kai and Liang, Jingyun and Cao, Jiezhang and Liu, Ce and Gong, Rui and Zhang, Yulun and Tang, Hao and Liu, Yun and Demandolx, Denis and others},
  booktitle={Proceedings of the IEEE/CVF Conference on Computer Vision and Pattern Recognition},
  pages={1775--1787},
  year={2023}
}

@inproceedings{barron2022mip,
  title={Mip-nerf 360: Unbounded anti-aliased neural radiance fields},
  author={Barron, Jonathan T and Mildenhall, Ben and Verbin, Dor and Srinivasan, Pratul P and Hedman, Peter},
  booktitle={Proceedings of the IEEE/CVF conference on computer vision and pattern recognition},
  pages={5470--5479},
  year={2022}
}

@inproceedings{blard2024overfitted,
  title={Overfitted image coding at reduced complexity},
  author={Blard, Th{\'e}ophile and Ladune, Th{\'e}o and Philippe, Pierrick and Clare, Gordon and Jiang, Xiaoran and D{\'e}forges, Olivier},
  booktitle={2024 32nd European Signal Processing Conference (EUSIPCO)},
  pages={927--931},
  year={2024},
  organization={IEEE}
}

@article{ha2016hypernetworks,
  title={HyperNetworks},
  author={Ha, David and Dai, Andrew and Le, Quoc V},
  journal={arXiv e-prints},
  pages={arXiv--1609},
  year={2016}
}

@inproceedings{schlag2017gated,
  title={Gated fast weights for on-the-fly neural program generation},
  author={Schlag, Imanol and Schmidhuber, J{\"u}rgen},
  booktitle={NIPS Metalearning Workshop},
  year={2017}
}

@inproceedings{schlag2021linear,
  title={Linear transformers are secretly fast weight programmers},
  author={Schlag, Imanol and Irie, Kazuki and Schmidhuber, J{\"u}rgen},
  booktitle={International conference on machine learning},
  pages={9355--9366},
  year={2021},
  organization={PMLR}
}

@article{volk2022example,
  title={Example-based hypernetworks for out-of-distribution generalization},
  author={Volk, Tomer and Ben-David, Eyal and Amosy, Ohad and Chechik, Gal and Reichart, Roi},
  journal={arXiv preprint arXiv:2203.14276},
  year={2022}
}

@inproceedings{sendera2023hypershot,
  title={Hypershot: Few-shot learning by kernel hypernetworks},
  author={Sendera, Marcin and Przewi{\k{e}}{\'z}likowski, Marcin and Karanowski, Konrad and Zi{\k{e}}ba, Maciej and Tabor, Jacek and Spurek, Przemys{\l}aw},
  booktitle={Proceedings of the IEEE/CVF winter conference on applications of computer vision},
  pages={2469--2478},
  year={2023}
}

@article{von2019continual,
  title={Continual learning with hypernetworks},
  author={Von Oswald, Johannes and Henning, Christian and Grewe, Benjamin F and Sacramento, Jo{\~a}o},
  journal={arXiv preprint arXiv:1906.00695},
  year={2019}
}

@inproceedings{chen2024mvsplat,
  title={Mvsplat: Efficient 3d gaussian splatting from sparse multi-view images},
  author={Chen, Yuedong and Xu, Haofei and Zheng, Chuanxia and Zhuang, Bohan and Pollefeys, Marc and Geiger, Andreas and Cham, Tat-Jen and Cai, Jianfei},
  booktitle={European conference on computer vision},
  pages={370--386},
  year={2024},
  organization={Springer}
}

@article{you2023generative,
  title={Generative neural fields by mixtures of neural implicit functions},
  author={You, Tackgeun and Kim, Mijeong and Kim, Jungtaek and Han, Bohyung},
  journal={Advances in Neural Information Processing Systems},
  volume={36},
  pages={20352--20370},
  year={2023}
}

@inproceedings{klocek2019hypernetwork,
  title={Hypernetwork functional image representation},
  author={Klocek, Sylwester and Maziarka, {\L}ukasz and Wo{\l}czyk, Maciej and Tabor, Jacek and Nowak, Jakub and {\'S}mieja, Marek},
  booktitle={International Conference on Artificial Neural Networks},
  pages={496--510},
  year={2019},
  organization={Springer}
}

@inproceedings{skorokhodov2021adversarial,
  title={Adversarial generation of continuous images},
  author={Skorokhodov, Ivan and Ignatyev, Savva and Elhoseiny, Mohamed},
  booktitle={Proceedings of the IEEE/CVF conference on computer vision and pattern recognition},
  pages={10753--10764},
  year={2021}
}

@inproceedings{catania2023nif,
  title={Nif: A fast implicit image compression with bottleneck layers and modulated sinusoidal activations},
  author={Catania, Lorenzo and Allegra, Dario},
  booktitle={Proceedings of the 31st ACM International Conference on Multimedia},
  pages={9022--9031},
  year={2023}
}

@inproceedings{li2026gaussianimage++,
  title={GaussianImage++: Boosted Image Representation and Compression with 2D Gaussian Splatting},
  author={Li, Tiantian and Zhang, Xinjie and Ge, Xingtong and Xu, Tongda and He, Dailan and Zhang, Jun and Wang, Yan},
  booktitle={Proceedings of the AAAI Conference on Artificial Intelligence},
  volume={40},
  number={8},
  pages={6442--6449},
  year={2026}
}

\appendix

\section{Fusing Modulations to Synthesis Net}
\label{sec:appendix:fusing}
In the encoding process, both the normalization and modulation parameters are fused into the parameters of $\text{MLP}^i$. The modulated network parameters subsequently included as part of the bitstream. Suppose $\boldsymbol{f}\in \mathbb{R}^{C\times H\times W}$  is the output feature of $\text{MLP}^s$
\begin{equation}
    \boldsymbol{f} = \text{MLP}^s(\hat{\boldsymbol{y}}_u),
\end{equation}
where $\hat{\boldsymbol{y}}_u$ is upsampled latents shown in Fig. 1 in main text. Note we omit batch dimensions for clarity. The modulated features is
\begin{equation}
    \bar{\boldsymbol{f}} = \frac{\boldsymbol{f} - \mu}{\sigma}, 
\end{equation}
\begin{equation}
    \tilde{\boldsymbol{f}} =\gamma\odot\bar{\boldsymbol{f}} + \beta,
\end{equation}
where $\mu=\mathrm{mean}(\boldsymbol{f}),\sigma=\mathrm{std}(\boldsymbol{f})$. Here we normalize $\boldsymbol{f}$ at $H$ and $W$ channel. Fraction represents element-wise division at $C$ dimension. $\odot$ is element-wise multiplication at $C$ dimension. $ \gamma \in \mathbb{R}^{C}$ and $\beta\in \mathbb{R}^{C}$ are modulation vectors.
The first layer of $\text{MLP}^i$ can be expanded as
\begin{align}
     \boldsymbol{f}' = &w^T\tilde{\boldsymbol{f}} + b \\
     = &w^T(\gamma\odot\bar{\boldsymbol{f}}) + w^T\beta + b \\
     = &\frac{w^T(\gamma\odot\boldsymbol{f})}{\sigma} - \frac{ w^T(\gamma\odot \mu) }{\sigma} + w^T\beta + b.
\end{align}

The net weight and bias included in bitstream are
\begin{equation}
    w' = \frac{\gamma^T\odot w}{\sigma^T},
\end{equation}
\begin{equation}
    b' = -\frac{ w^T(\gamma\odot \mu) }{\sigma} + w^T\beta + b.
\end{equation}

In decoding phase, only $w'$ and $b'$ are required, with no need for normalization and modulation. In our experiments, we use first 3 layer as $\text{MLP}^s$ and last layer as $\text{MLP}^i$.

\section{More Implementation Detatils}

\subsection{Fast Chunking Algorithms}

Algo. \ref{alg:expand_symbols_zigzag} is a fast chunking algorithm that splits the symbol stream based on end-of-block markers (eob\_marker) and expands the zero symbols according to the pRLE code table. 
This function requires precomputing the offsets of all eob\_marker to enable parallel processing. We use thrust::copy\_if to efficiently obtain all offsets. For synthesis network $g_s$, we fuse all layers in a single CUDA kernel to achieve efficient processing. All matrix multiplications in $g_s$ are implemented using APIs provided by nvcuda::wmma.
\begin{algorithm}
\caption{Expand Symbols}
\label{alg:expand_symbols_zigzag}
\begin{algorithmic}[1]
\Procedure{expand\_symbols}{symbols, offsets, output}
    
    \State offset\_start $\gets$ 0
    \If{gid $\neq$ 0} \Comment{gid is CUDA thread id}
        \State offset\_start $\gets$ offsets[gid$-1$] $+$ 1
    \EndIf
    \State offset\_end $\gets$ offsets[gid]
    
    \State eob\_marker $\gets$ symbols[offset\_end] \Comment{Get eob\_marker from symbol stream}
    \State output\_symbols $\gets$ output $+$ gid $\times$ 64
    
    \State offset $\gets$ 0
    \For{$i \gets$ offset\_start \textbf{to} offset\_end$-1$}
        \State $s \gets$ symbols[$i$]
        \If{$s \geq$ eob\_marker $-$ 17 $+$ 1} \Comment{If $s$ is a symbol in the pRLE code table}
            \For{$j \gets 0$ \textbf{to} $s + 17 -$ eob\_marker $-1$} 
                \State output\_symbols[ZIGZAG\_ORDER[offset]] $\gets 0$
                \State offset $\gets$ offset $+$ 1
            \EndFor
        \Else
            \State output\_symbols[ZIGZAG\_ORDER[offset]] $\gets$ symbols[$i$]
            \State offset $\gets$ offset $+$ 1
        \EndIf
    \EndFor
\EndProcedure
\end{algorithmic}
\end{algorithm}


\subsection{Details of Baseline Methods}
For COIN~\cite{dupont2021coin} and GaussianImage~\cite{zhang2024gaussianimage}, We reproduce all the results using the original source code. It should be noted that although our method only used MSE as the distortion metric during the training, we still adhered to the original loss function settings for the baseline methods, even if the original approaches employed loss functions other than simple MSE. Additionally, for all methods, we did not use model versions specifically trained for different test metrics. For instance, in the case of Factorized model~\cite{balle2017end}, all experiments utilized a pre-trained version optimized with MSE as the loss function.

\subsection{Evaluation Protocol}
Due to the fact that speed evaluation is susceptible to interference from various factors, including hardware occupancy and PyTorch's inherent asynchronous design, we have carefully designed the speed evaluation process and ensured exclusive access to the hardware during assessment. 
Although there are currently many implementations of JPEG, in order to achieve a fair comparision, it is necessary to use a GPU-optimized version. However, since nvNVJPE is not open-source, we implemented our binding based on previous implemntation \footnote{https://github.com/UsingNet/nvjpeg-python}. 
We have also meticulously optimized the input and output components of nvJPEG to minimize any additional overhead as much as possible. Besieds, we run encoding and decoding in same process without saving bitstream to disk for nvJPEG, which is intended to eliminate the I/O time. In experiment, we find running encoding and decoding in same process is sufficient to warmup nvJPEG. For all method, we report the result of serial decoding all images in corresponding dataset, following is part of benchmark code for nvJPEG

\begin{lstlisting}[float, floatplacement=h,
  caption={Snippet used to mesure nvJPEG decoding speed.},
  label={lst:nvjpeg}]
for idx in range(len(bs_pack)):
    jpeg_bytes, img, f = bs_pack[idx]  # encoded bitstream
    torch.cuda.synchronize()
    start_time = time.time()
    img_decoded = coder.decode(jpeg_bytes) # run nvJPEG decoder
    torch.cuda.synchronize()
    end_time = time.time()
    total += end_time - start_time
print('decoding time', total / len(bs_pack))
\end{lstlisting}

\section{More Experiment Results}






\subsection{Extended Results}
 \begin{figure}[t]
    \center
\includegraphics[width = 0.98\textwidth, trim=80 0 100 0,clip]{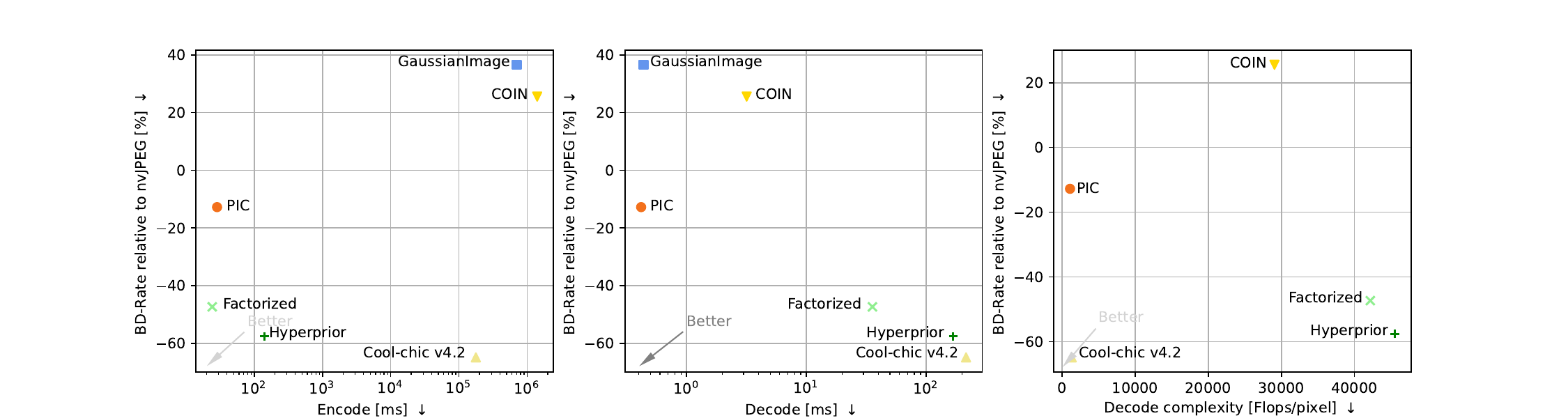}


  \caption{Results of comprehensive comparesion for both RD performance and practical performance on Kodak dataset.}
    \label{fig:ablation:rd_speed}
\end{figure}

\cref{fig:ablation:rd_speed} uses BD-rate as the vertical axis to compare the encoding/decoding speed and decoding complexity of different methods. While our method does not achieve optimal performance across all dimensions, it demonstrates balanced capabilities with no significant weaknesses in the three metrics. In contrast, autoencoder methods~\cite{balle2017end,balle2018variational} offer fast encoding capabilities but suffer from high complexity, whereas Cool-chic~\cite{ladune2022cool,leguay2023low} maintains low complexity at the cost of significantly slower encoding and decoding speeds.

\subsection{Latents Visualizations}
\label{sec:app:latent}
\begin{figure}[t]
    \center
  \subfloat[$\lambda = 0.001$]{\includegraphics[width = \textwidth, trim=160 0 130 0,clip]{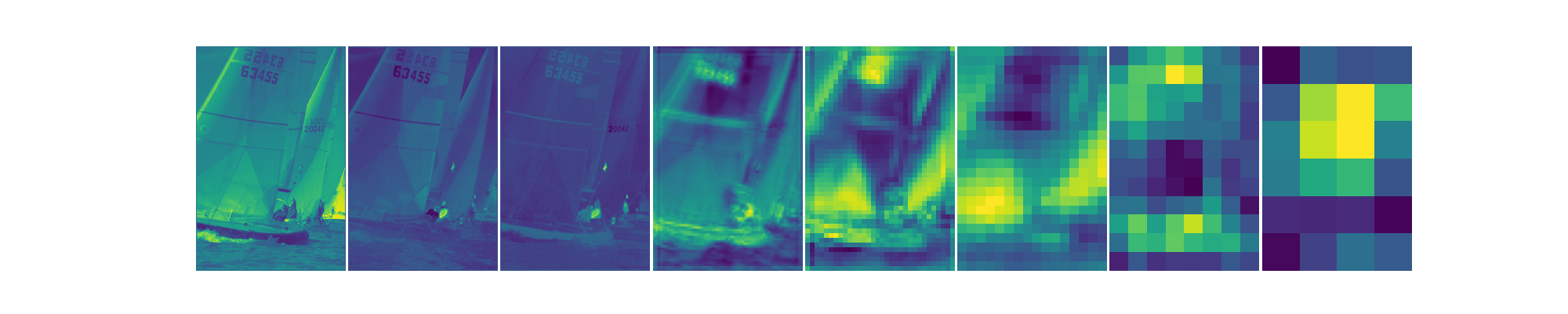}}

  \subfloat[$\lambda = 0.01$]{\includegraphics[width = \textwidth, trim=160 0 130 0,clip]{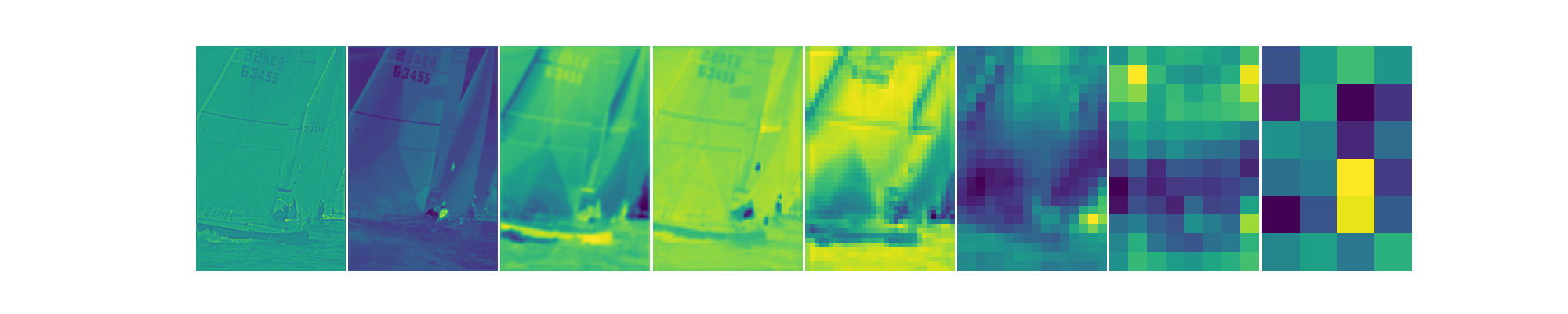}}

  \caption{Visualizations of latents. These figures demonstrate unnormalized value for each latent.}
    \label{fig:supp_latent}
\end{figure}

\begin{figure}[t]
    \center
  \subfloat[$\lambda = 0.001$]{\includegraphics[width = \textwidth, trim=160 0 130 0,clip]{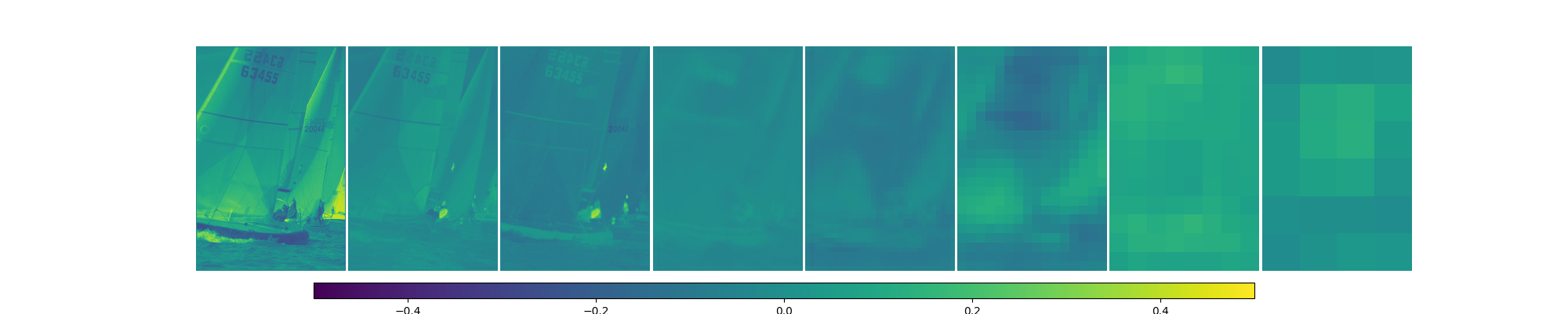}}

  \subfloat[$\lambda = 0.01$]{\includegraphics[width = \textwidth, trim=160 0 130 0,clip]{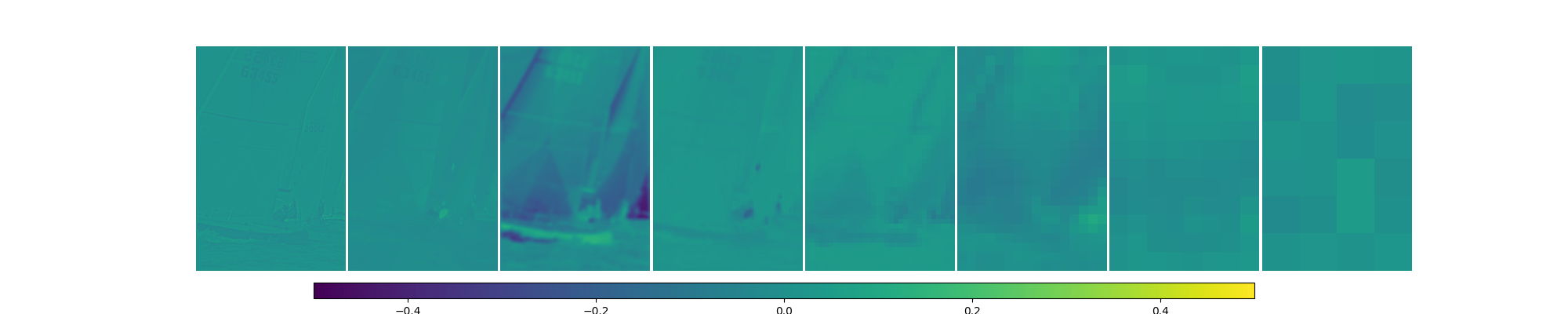}}

  \caption{Visualizations of latents. These figures demonstrate normalized value for each latent.}
    \label{fig:supp_latent_norm}
\end{figure}

To better demonstrate the model's performance under different bitrates, we visualized the latent representations when $\lambda=\{0.01, 0.001\}$ . \cref{fig:supp_latent} shows the visualizations without normalization. Overall, the latents at different resolutions reflect varying levels of detail from the original image. \cref{fig:supp_latent_norm} presents the results after applying same normalization to all latents. Although these latents do not directly indicate the magnitude of the corresponding symbols, they indirectly reveal that at smaller $\lambda$, higher-resolution latents retain more details, while at bigger $\lambda$, more details are contained in relatively lower-resolution latents. This also reflects the bits allocation tendencies under different settings.


\subsection{Ablation study of model architecture}
\label{sec:app:abl}

\begin{table}[t]

\centering
\caption{Ablation result of RLE. RLE on 0 achieves the best performance among four settings. BD-Rate is calculated relative to RLE on 0.}
\label{abl_rle}
\begin{tabular}{lc}
    \toprule
    Setting	& BD-Rate($\downarrow$) \\
    \midrule
    No RLE          & 150.68\% \\
    RLE on 0        & \textbf{0\%} \\
    RLE on 0, 1     & 1.18\% \\
    RLE on -1, 0, 1	& 2.70\%  \\
    \bottomrule
    \end{tabular}
\end{table}

\begin{table}[t]
\centering
\caption{Start means we place $\mathrm{MLP}^i$ at the beginning of the synthesis network. Middle means we place an $\mathrm{MLP}^i$ between two shared MLPs. BD-Rate is calculated relative to default configuration, in which $\mathrm{MLP}^i$ are placed at the end of synthesis network.}
\label{loc_mlp}
\begin{tabular}{lc}
        \toprule
Location & BD-Rate ($\downarrow$) \\
    \midrule
Start & 21.46\% \\
Middle & 2.43\% \\
End (default for PIC) & \textbf{0\%} \\
    \bottomrule
\end{tabular}

\end{table}

Given implementation complexity considerations, we opted for a simplified RLE. Because our symbol range is determined by training, we cannot fix an RLE encoding table like JPEG, but dynamically determine it through the transformed symbol stream. \cref{abl_rle} presents a performance comparison of performing RLE encoding on more symbols (-1 and 1). From the results in the table, it can be seen that performing RLE only on 0 is not only easier to implement in engineering, but can also achieve better results in certain situations. 

Another ablation study explored the position of  $\mathrm{MLP}^i$. We investigated placing the $\mathrm{MLP}^i$ at the input layer, intermediate layer, and output layer of the synthesis network, respectively. \cref{loc_mlp} presents the result. We found placing $\mathrm{MLP}^i$ at the end of synthesis network will obtain the best performance.


Furthermore, the width of the synthesis network is also a parameter that needs to be examined. \cref{abl_width} displays the performance under different widths. Overall, the configuration with a width of 16 is the optimal choice, taking into account decoding speed, RD performance, and hardware compatibility.

\begin{table}[t]

\centering
\caption{Ablation result of the synthesis network width. BD-Rate is calculated relative to width 16.}
\label{abl_width}
\begin{tabular}{lc}
    \toprule
    Width	& BD-Rate($\downarrow$) \\
    \midrule
    8         & 3.03\% \\
    16        & \textbf{0\%} \\
    32     & 5.84\% \\
    \bottomrule
    \end{tabular}

\end{table}




\subsection{Ablation of modulation }
\label{sec:ablation:mod_nocc}

 \begin{figure}[ht]
    \center

  \includegraphics[width = 0.6\textwidth]{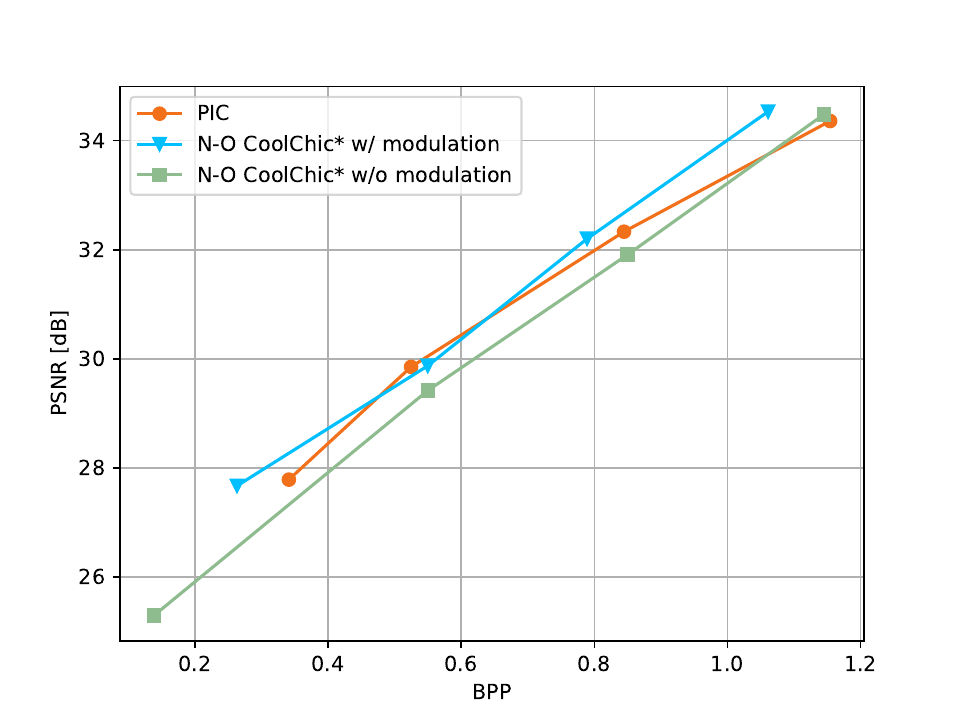}

  \caption{Ablation results of modulation mechanism on N-O Cool-Chic-like (mark by *) architecture~\cite{blard2024overfitted}. It should also be noted that N-O Cool-Chic is not open-source, so the results presented here are only a preliminary reproduction without hyperparameters and training recipts tuning and may differ from those reported in the original paper. }
    \label{fig:ablation:nocc_ab}
\end{figure}

In addition to exploring the effect of the modulation mechanism on the model architecture proposed in this paper, we also validated the effectiveness of this mechanism on model architectures similar to N-O Cool-Chic. \cref{fig:ablation:nocc_ab} illustrates the results of the ablation results. Since the last two layers of the N-O Cool-Chic's synthesis network are convolutional layers, we applied modulation to the features between the first two linear layers. The results show the effectiveness of the modulation mechanism in this architecture as well. Moreover, N-O Cool-Chic* with modulation has surpassed the method proposed in this paper in terms of RD performance, highlighting the potential of the end-to-end INR paradigm for further advancements in its network architecture and training recipts.

\subsection{Ablation of bitstream }

\begin{figure}[ht]
    \center

  \includegraphics[width = 0.6\textwidth]{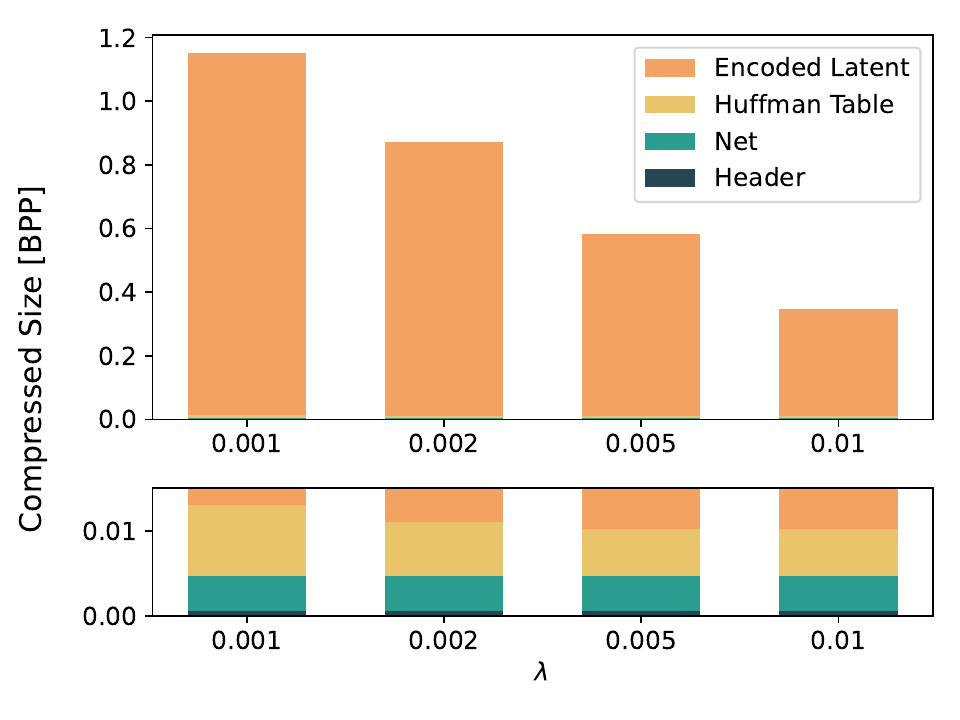}

  \caption{Bitstream breakdown. }
    \label{fig:ablation:breakdown_ab}
\end{figure}

The \cref{fig:ablation:breakdown_ab} shows the proportion of different components in the encoded bitstream on the Kodak dataset. The encoded latent variables account for the vast majority of the bitstream, and the codebook size increases with the bitstream but has little impact on the overall encoded size.

\end{document}